\documentclass[11pt]{article}
\usepackage[margin=1in]{geometry}
\usepackage{amsmath,amssymb,amsthm,mathtools}
\usepackage{booktabs,array,longtable,multirow}
\usepackage{graphicx}
\usepackage[numbers,sort&compress]{natbib}
\usepackage[hidelinks]{hyperref}
\usepackage{xcolor}

\theoremstyle{definition}

\theoremstyle{remark}

\newcommand{\E}{\mathbb{E}}

\newcommand{\kv}{\mathrm{kv}}
\newcommand{\kmax}{k_{\max}}
\newcommand{\rbar}{\bar\rho}

\newcommand{\Vs}{V_{\mathrm{slot}}}
\newcommand{\Vu}{V_{\mathrm{dedup}}}
\newcommand{\Es}{E_{\mathrm{slot}}}
\newcommand{\Eu}{E_{\mathrm{dedup}}}

\setcitestyle{authoryear,round,citesep={;},aysep={,},yysep={,}}
\usepackage{graphicx}
\usepackage{placeins}

\title{Speculative Decoding for Mixture-of-Experts Models:\\ Expert Bytes, Kernel Bandwidth and the Best Draft Length}
\author{Yuzheng Lan\\ \small ReClan AI Studio, LLC\\ \small\texttt{research@reclan.ai}}
\date{}

\begin{document}
\maketitle

\begin{abstract}
Speculative decoding checks several drafted tokens in one verification pass. On a mixture-of-experts (MoE) model, consecutive tokens often select the same experts. A deduplicating engine reads each selected expert once per pass, so it requests fewer bytes than a per-slot engine, which reads it once per token. We write a step's time as the bytes each kernel requests over its effective bandwidth, plus fixed time. Under geometric acceptance, we derive from this step time when speculation is faster than plain decoding. The best draft length is a monotone staircase in the acceptance rate. We bound how a bandwidth error moves each result, and we check the main steps in Lean. Deduplication is faster exactly where its share $\bar u(n)$ of the per-slot bytes is below its ratio $r(n)$ of the per-slot bandwidth. On llama.cpp Metal, MLX-LM and llama.cpp CUDA, which we modified to deduplicate, the per-slot kernel has the higher effective bandwidth ($r=0.32$ to 0.75). Thus deduplication is faster only from 5 to 8 tokens per pass, or never. We fit the bandwidths on one layer's expert operation, run alone, in under an hour per model. Bytes from routing traces over these bandwidths predict the time that deduplication saves. The measured saving lies inside its error bound at every point on CUDA and MLX-LM and at 36 of 48 points on Metal. The crossover is the first pass size at which deduplication is faster. The prediction places it within one pass size in 10 of 12 configurations of engine, model and depth. We computed these predictions after the measurements. On DeepSeek-V2-Lite, operation times measured before its speculative runs give the best draft length on CUDA within one at all 22 acceptance values.
\end{abstract}

\section{Introduction}
\label{sec:intro}

A mixture-of-experts (MoE) layer replaces one feed-forward block by $N$ experts and a router that sends each token to $m$ of them \citep{shazeer2017,arXiv:2101.03961}. At batch size one, memory reads limit decoding \citep{arXiv:2211.05102,arXiv:2607.17283,arXiv:2609.04238}. In speculative decoding \citep{arXiv:2211.17192,arXiv:2302.01318}, a small draft model proposes $k$ tokens, and this number is the \emph{draft length}. The \emph{target}, the model being decoded, checks these tokens with the current token in one \emph{verification pass}. The \emph{acceptance rate} $a$ is the probability that a drafted token survives. On an MoE model, the pass needs the experts of all $k+1$ verified tokens. Published measurements disagree on whether speculation then speeds up decoding \citep{arXiv:2506.20675,arXiv:2610.15417}. Consecutive tokens often choose the same experts \citep{arXiv:2606.21868,arXiv:2608.18261}. An \emph{engine} is the inference software that runs the model. A \emph{kernel} is a GPU routine that the engine launches for one operation. The kernel that matters here runs the \emph{expert operation}. This operation is the set of matrix products (gate, up and down projections) of each token of a verification pass with the weights of its selected experts. A \emph{slot} is one of a token's $m$ expert selections, so a pass of $n$ tokens has $nm$ slots. A \emph{per-slot} kernel multiplies each token by each of its selected experts separately, with one expert read per slot. A \emph{deduplicating} kernel groups the pass's tokens by expert and reads each distinct expert once. We call an engine per-slot or deduplicating by the kernel that it runs on verification passes. So reuse lowers the bytes of a pass only on a deduplicating engine.

Consider an affine step time $1+\chi k$, with $\chi$ the cost of a drafted token relative to a target step. For this step time, \citet{arXiv:2211.17192} show that some draft length is faster than plain decoding exactly when $a>\chi$. They also give the first \emph{step point}, the acceptance at which the best draft length first exceeds zero. Section~\ref{sec:spec-theory} extends both results to any step time. It also gives every step point when each further draft adds at least as much time as the one before. The cost of an MoE verification pass is not affine in $k$. Its bytes depend on how many distinct experts the verified tokens select. Its time depends on how fast each kernel moves those bytes.

Prior systems decide whether to speculate either from measurement or from a model of the verification cost. \citet{arXiv:2506.20675} disables speculation when a utility measured online falls below one. Among cost models for MoE targets, MoESD \citep{arXiv:2505.19645} derived the expected number of distinct experts that $n$ tokens select under uniform routing: $N(1-q^n)$ with $q=1-m/N$. \citet{arXiv:2603.09983} assumes deduplication and no fixed cost. Its speedup condition is on the ratio of the distinct experts that the tokens of a verification pass select to those of one token. So it is a condition on how often consecutive tokens repeat experts. S2-MoE \citep{arXiv:2608.15018} charges a verification the newly introduced experts times a per-expert latency. SpecFit \citep{arXiv:2610.15417} predicts cycle time as an intercept plus a coefficient times the per-layer union of experts. Both charge every expert read the same time at every pass size, which is a single bandwidth for every kernel. Neither derives the \emph{staircase}, the best draft length as a function of acceptance. Neither derives a speedup from reuse together with the bandwidth of each kernel.

Next-token expert reuse has been measured at 44 to 48\% on Qwen3-30B-A3B \citep{arXiv:2606.21868} and 0.444 in llama.cpp router traces \citep{arXiv:2608.18261}. MonoMoE \citep{arXiv:2609.04244} finds that token-major kernels that group tokens by expert underuse memory bandwidth when few tokens reach each expert. It replaces them by a fused kernel that loops over expert weights and applies each to all of its tokens. Opportunistic Expert Activation \citep{arXiv:2511.02237} re-routes tokens to experts that a batch already loads. Neither models when a deduplicating kernel is faster than a per-slot one. Engine developers tune that switch by hand \citep{llamacpp30047}. An MLX issue reports that the deduplicating product at one row runs near half its bandwidth at four rows \citep{mlx4509}. Both are engineering reports without peer review.

This paper separates the two inputs of a step's time. A speedup is a ratio of step times. A step's time is the bytes that its kernels request over the rate at which they move them, plus fixed time. That rate is the \emph{effective bandwidth} of each kernel. The bytes follow from routing alone. On CUDA, they match a hardware count of DRAM bytes. The bandwidths depend on the kernel: on every engine and model that we measured, the per-slot kernel has the higher effective bandwidth. On a pass of $n$ tokens, deduplication requests a share $\bar u(n)$ of the per-slot bytes, and routing fixes this share. Deduplication moves these bytes at a ratio $r(n)$ of the per-slot bandwidth, and the kernels fix this ratio. So deduplication is faster exactly where $\bar u(n)<r(n)$. Its \emph{saving}, the time it removes from a pass, is the per-slot expert time times $1-\bar u(n)/r(n)$. Section~\ref{sec:motivation} measures the bytes and the bandwidths. Section~\ref{sec:spec-theory} works in that step time. It gives the condition under which a draft length is faster than plain decoding, and the staircase. On a deduplicating engine, it gives the \emph{reuse threshold}, the least mean reuse at which a draft length is faster, and the speedup. It also gives bounds on how a bandwidth error moves each prediction. These results are proved in Lean~4 \citep{lean4} against Mathlib \citep{mathlib2020}. Section~\ref{sec:dedup} tests them on deduplicating versions of llama.cpp Metal, MLX-LM and llama.cpp CUDA. We measure the bandwidth on the expert operation, in under an hour per model. Bytes from routing over this bandwidth predict the saving within its stated error on CUDA and MLX-LM. They predict the crossover within one pass size in 10 of 12 configurations. A \emph{configuration} is an engine, a model and a context depth. On DeepSeek-V2-Lite, we measured operation times before any of its speculative runs. These times predict the best draft length on CUDA within one at every acceptance value. In contrast, the prediction from fits run afterwards misses its CUDA crossover. On Metal, the verify time needs one calibration of the time outside the expert operation.
\label{sec:setting}

\paragraph{Routing, union and reuse.} A decoder has $L$ MoE layers of $N$ experts. In layer $l$, the router selects a set $S_{l,t}$ of $m$ experts, $1\le m<N$, for the token at position $t$. We call a law on the array $(S_{l,t})$ a \emph{routing process}. It is \emph{stationary} when a window of $n$ tokens has the same law wherever it starts. A pass over $n$ consecutive tokens needs their $X_{l,t}(n)=|S_{l,t}\cup\dots\cup S_{l,t+n-1}|$ distinct experts, with mean $U_l(n)$, $U_l(0)=0$. We define the \emph{consecutive-token reuse}
\begin{equation}
\rho_l(n)=\frac{1}{m}\,\E\bigl|S_{l,t+n-1}\cap(S_{l,t}\cup\dots\cup S_{l,t+n-2})\bigr|,\qquad\rho_l(1)=0,
\label{eq:reuse}
\end{equation}
the expected share of a token's experts that one of the $n-1$ tokens before it also selected. At $n=2$ it is the next-token reuse measured by \citet{arXiv:2606.21868}. Each token therefore adds $m(1-\rho_l(n))$ new experts on average. By linearity of expectation, for every stationary process,
\begin{equation}
U_l(n)=m\sum_{j=1}^{n}\bigl(1-\rho_l(j)\bigr).
\label{eq:union-reuse}
\end{equation}
Under independent uniform routing $1-\rho_l(n)=q^{\,n-1}$ with $q=1-m/N$ \citep{arXiv:2505.19645}.

\paragraph{Bytes.} Let $w_l$ be the bytes of one expert of layer $l$, $W=\sum_l w_l$, and $D$ the bytes read once per pass whatever the routing. A \emph{trace} is the recorded expert sets of one token sequence. On every trace, a deduplicating kernel requests $\sum_l w_l X_{l,t}(n)$ expert bytes for a pass of $n$ tokens. A per-slot kernel requests $mnW$, one request per slot. In expectation, the \emph{union bytes} are $\Eu(n)=\sum_l w_l U_l(n)$, and $\Es(n)=mnW$. So a per-slot kernel requests
\begin{equation}
\Es(n)-\Eu(n)=m\sum_l w_l\sum_{j=1}^{n}\rho_l(j)\ \ge\ 0
\label{eq:class-gap}
\end{equation}
more bytes. These are the bytes that reuse would save. They are requested bytes, and on-chip caches can serve some requests.

\paragraph{Speculative decoding.} A step drafts $k\in\{0,\dots,\kmax\}$ tokens, with $\kmax$ the largest draft length allowed. It verifies them with the current token in one pass of $k+1$ tokens. The case $k=0$ is plain decoding. With kernel $p\in\{\mathrm{slot},\mathrm{dedup}\}$ the step requests $D+E_p(k+1)+\kv(k)$ bytes, with $\kv(k)\ge0$ the key-value cache bytes it reads \citep{arXiv:2211.05102}. Under \emph{geometric acceptance} at rate $a\in[0,1)$ the drafts are accepted in order, each with probability $a$, until the first rejection. A step then emits $G_a(k)=1+a+\dots+a^k$ tokens on average \citep{arXiv:2211.17192}. We generalize the walltime analysis of \citet{arXiv:2211.17192}. The throughput is $T_a(k)=G_a(k)/V(k)$ tokens per second. Here $V(k)$ is the \emph{step time}, the wall-clock time of one step that drafts $k$ tokens and verifies $k+1$ (Section~\ref{sec:spec-affine} gives its form). The \emph{best draft length} $k^*(a)$ is the smallest maximizer of $T_a$. So the staircase is $a\mapsto k^*(a)$. Speculation \emph{gives a speedup} at $a$ when $k^*(a)\ge1$. The $j$-th step point $a_j$ is the infimum of the acceptances $a\in[0,1)$ at which $k^*(a)\ge j$. It is taken as 1 when there is none. Finally, $c_i=V(i)-V(i-1)$ is the time that the $i$-th draft adds. Appendix~\ref{app:notation} lists every symbol.
\section{Setup}
\label{sec:motivation}

This section measures the two inputs separately: the bytes and the effective bandwidths. The measurements use an Apple M4 Max (546 GB/s, \citealp{applem4max}) with llama.cpp on Metal and MLX-LM. They also use an NVIDIA RTX 4070 Ti SUPER (672 GB/s, 48 MiB L2 cache) with llama.cpp on CUDA. The models are OLMoE-1B-7B-0125 \citep{olmoe2025} ($N=64$, $m=8$), Qwen3-30B-A3B \citep{qwen3} ($N=128$, $m=8$), gpt-oss-20b \citep{gptoss2025} ($N=32$, $m=4$) and DeepSeek-V2-Lite (Section~\ref{sec:dedup-blind}). Appendix~\ref{sec:spec-tests} gives the setup of the first three. Byte units are decimal (1~GB is $10^9$ bytes), except MiB, which is $2^{20}$ bytes.

\begin{figure}[t]
\centering

\includegraphics[width=0.45\linewidth]{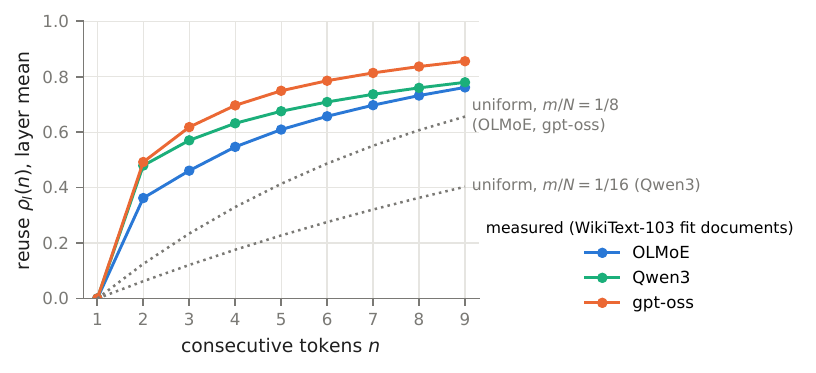}
\caption{Consecutive-token reuse $\rho_l(n)$, layer mean, on the WikiText-103 fit documents, against the number $n$ of consecutive tokens. Dotted: independent uniform routing, $1-q^{\,n-1}$ \citep{arXiv:2505.19645}. In every figure, color marks the model: blue OLMoE, green Qwen3, orange gpt-oss and brown DeepSeek-V2-Lite.}
\label{fig:reuse}
\end{figure}

\paragraph{Expert reuse.} We record the experts each token selects on the 60 articles of the WikiText-103 test set \citep{wikitext2016}. A hash of the document identifier selects 32 of them for fitting. On these, $\hat\rho_l(2)$ is the estimate of $\rho_l(2)$ in~\eqref{eq:reuse}, and Appendix~\ref{sec:spec-tests} gives the estimator. Its layer mean is 0.363, 0.480 and 0.492 for OLMoE, Qwen3 and gpt-oss. Uniform routing gives $m/N=0.125$, 0.0625 and 0.125. At $n=9$ the layer mean is 0.762, 0.780 and 0.856 (Figure~\ref{fig:reuse}). Yet on the short passes of speculation (up to 8 tokens), llama.cpp Metal, MLX-LM and llama.cpp CUDA multiply each token's experts separately (Appendix~\ref{sec:engines}). So each draft adds the full expert bytes of one more token. With a real draft model on llama.cpp Metal, plain decoding is fastest on 33 of 37 held-out MT-Bench prompts \citep{mtbench2023} (Appendix~\ref{sec:spec-tests-real}).

\begin{figure}[t]
\centering

\includegraphics[width=\linewidth]{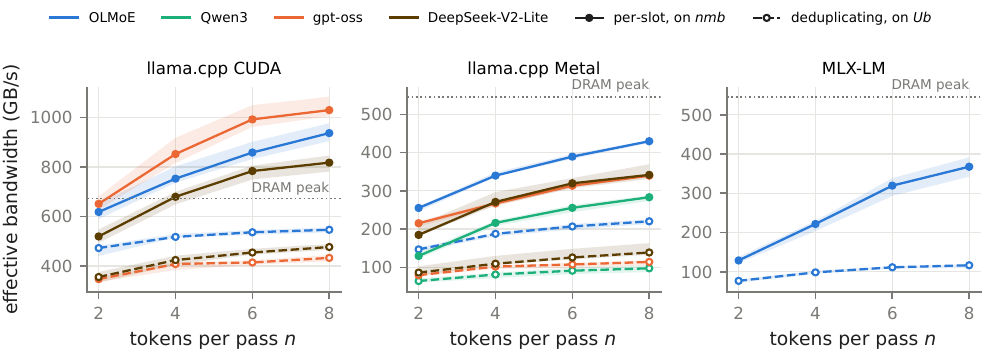}
\caption{Effective bandwidth of each kernel on the expert operation alone. Per-slot kernel: on requested bytes $nmb$ (solid line, filled markers). Deduplicating kernel: on union bytes $Ub$ (dashed line, open markers). Median and interquartile range over all MoE layers and windows of recorded routing; color: model; dotted: nominal DRAM bandwidth. CUDA was measured on OLMoE, gpt-oss and DeepSeek-V2-Lite, and MLX-LM on OLMoE.}
\label{fig:beta}
\end{figure}

\paragraph{Effective bandwidth.} We write $\beta_p(n)$ for the effective bandwidth of kernel $p$ on a pass of $n$ tokens: its requested expert bytes over its time. We measure it in a separate benchmark of the expert operation of one MoE layer. The benchmark runs each engine's own kernels, with no other part of the model. It takes the expert ids of each pass as input, and we give it two kinds of ids. In the first kind, we choose the ids ourselves. For each pass size $n$, we set how many distinct experts $U$ the $nm$ slots of the pass name. The range goes from heavy repetition to none. This lets a fit separate the time of the first read of an expert in a pass from the time of a repeated read. In the second kind, each pass copies its ids from a window of consecutive tokens of \emph{recorded routing}, here the WikiText-103 traces. So the kernel sees the routing that the model produces. We write $r(n)=\beta_{\mathrm{dedup}}(n)/\beta_{\mathrm{slot}}(n)$ for the bandwidth ratio. We write $\bar u(n)=\Eu(n)/\Es(n)$ for the share of the per-slot bytes that deduplication requests. A kernel whose repeated reads hit an on-chip cache has a higher effective bandwidth than one that reads every requested byte from memory. The per-slot kernel moves its requested bytes at a higher effective bandwidth than the deduplicating kernel moves its distinct ones (Figure~\ref{fig:beta}). This holds on every engine and model measured, at every $n$ from 2 to 8. The ratio $r(n)$ is 0.37 to 0.75 at $n=2$ and 0.32 to 0.59 at $n=8$. It is lower at $n=8$ than at $n=2$ in all eight engine and model pairs (Table~\ref{tab:beta}). On the passes with chosen ids, $U$ of the $nm$ slots are the first read of their expert in the pass, and $nm-U$ repeat one. We fit the time of the per-slot kernel as
\begin{equation}
t_{\mathrm{slot}}=t_{\mathrm{op}}+c_{\mathrm{u}}\,Ub+c_{\mathrm{re}}\,(nm-U)\,b,
\label{eq:reread}
\end{equation}
where $t_{\mathrm{slot}}$ is the median time, over repetitions, of the expert operation of one layer with the per-slot kernel. Here $t_{\mathrm{op}}$ is the part of that time that does not grow with the bytes, such as kernel launches. The bytes of one expert of the timed layer $l$ are $b=w_l$. The constant $c_{\mathrm{u}}$ is the time per byte of a first read, and $c_{\mathrm{re}}$ is the time per byte of a repeated slot. We fit the constants $t_{\mathrm{op}}$, $c_{\mathrm{u}}$ and $c_{\mathrm{re}}$ per layer by least squares, over the passes with chosen ids at $n=2,\dots,9$. The \emph{re-read cost} $c_{\mathrm{re}}/c_{\mathrm{u}}$ is 1 when a repeated slot is read again from memory and near 0 when an on-chip cache serves it. On CUDA and MLX-LM, caching explains part of the gap. The re-read cost is 0.18 to 0.34 on CUDA and 0.27 to 0.40 on MLX-LM (Table~\ref{tab:beta}). On CUDA, the median per-slot bandwidth also exceeds the DRAM bandwidth of the card, 672 GB/s, from $n=4$ (Figure~\ref{fig:beta}). On llama.cpp Metal the re-read cost is 0.88 to 1.04, so caching does not explain the gap there. There the gap lies in the deduplicating kernel. At $n=2$ it moves 65 to 147 GB/s, against 131 to 250 GB/s for the per-slot kernel.

\begin{table}[t]\centering\small

\caption{Effective bandwidth of each kernel in GB/s, over all MoE layers and windows of recorded routing, including the fixed work of the operation. The values of $r$ and $\bar u$ come from bytes and times summed over layers and windows. Re-read cost: $c_{\mathrm{re}}/c_{\mathrm{u}}$ of~\eqref{eq:reread}, as a range over two layers ($-$: no fit on chosen ids). The deduplicating kernel is faster where $\bar u<r$~\eqref{eq:grp-faster}.}
\label{tab:beta}
\renewcommand{\arraystretch}{1.12}
\begin{tabular}{llc cccc cccc}
\toprule
& & re-read & \multicolumn{4}{c}{$n=2$} & \multicolumn{4}{c}{$n=8$} \\
\cmidrule(lr){4-7}\cmidrule(lr){8-11}
engine & model & cost & $\beta_{\mathrm{slot}}$ & $\beta_{\mathrm{dedup}}$ & $r$ & $\bar u$ & $\beta_{\mathrm{slot}}$ & $\beta_{\mathrm{dedup}}$ & $r$ & $\bar u$ \\
\midrule
llama.cpp CUDA & OLMoE & 0.18-0.22 & 618 & 464 & 0.75 & 0.82 & 933 & 546 & 0.58 & 0.50 \\
llama.cpp CUDA & gpt-oss & 0.34 & 654 & 346 & 0.53 & 0.72 & 1018 & 434 & 0.43 & 0.36 \\
llama.cpp CUDA & DeepSeek-V2-Lite & $-$ & 523 & 355 & 0.68 & 0.83 & 814 & 480 & 0.59 & 0.62 \\
llama.cpp Metal & OLMoE & 0.88-0.90 & 250 & 147 & 0.59 & 0.82 & 428 & 219 & 0.51 & 0.50 \\
llama.cpp Metal & Qwen3 & 0.92-0.98 & 131 & 65 & 0.50 & 0.77 & 280 & 98 & 0.35 & 0.43 \\
llama.cpp Metal & gpt-oss & 0.98-1.04 & 213 & 79 & 0.37 & 0.72 & 339 & 114 & 0.34 & 0.36 \\
llama.cpp Metal & DeepSeek-V2-Lite & $-$ & 196 & 91 & 0.47 & 0.82 & 350 & 148 & 0.42 & 0.62 \\
MLX-LM & OLMoE & 0.27-0.40 & 129 & 75 & 0.58 & 0.82 & 358 & 116 & 0.32 & 0.50 \\
\bottomrule
\end{tabular}
\end{table}

\paragraph{Bytes against bandwidth.} A deduplicating engine requests the share $\bar u(n)$ of the per-slot bytes, at the lower bandwidth. So, with equal fixed work, it is faster exactly where $\bar u(n)<r(n)$~\eqref{eq:grp-faster}. On the four models, $\bar u(2)$ is 0.72 to 0.83 (Table~\ref{tab:beta}), above $r(2)$ in every engine and model pair. So at two tokens the deduplicating kernel is slower everywhere. The bytes that reuse saves become time saved only from a pass size that the bandwidth ratio sets (Section~\ref{sec:dedup-cross}). Predicting the speedup needs both inputs: the bytes, which routing fixes, and the bandwidth of each kernel, which the kernels fix.

\paragraph{Testing the inputs.} A bandwidth is bytes over time. So a wrong byte count would be absorbed into a bandwidth fitted on the same passes. For that reason, this paper uses no timing as evidence for a byte count. The fit~\eqref{eq:reread} only locates the source of the higher bandwidth of the per-slot kernel. The bytes come from the routing and from the source code of the kernels. On CUDA, a hardware counter reads the DRAM bytes that each kernel moves, with no timing involved, and confirms them (Section~\ref{sec:dedup-bytes}). The Apple engines expose no such counter, so there the bytes rest on the source reading. The bandwidths come from the benchmark of the expert operation alone. Their stated error $\varepsilon$ is a bound on the relative error of a measured bandwidth, and it comes from a split inside the benchmark. We make the linear fits of operation time to bytes on the passes with chosen ids. These fits are~\eqref{eq:reread} for the per-slot kernel and its counterpart for the deduplicating kernel (Appendix~\ref{app:dedup-evidence}). We then score them on the windows of recorded routing, which the fits do not see. The test of the two inputs together is prediction. Bytes from routing over bandwidths from the benchmark give whole-model \emph{verify times} $\tau_p(n)$, the time of one verification pass. They also give savings, best draft lengths and \emph{crossovers}. A crossover is the smallest pass size from which the deduplicating kernel is faster. For a measurement, it is the smallest pass size from which the point estimate of the measured saving stays positive. We compare these predictions with whole-model runs that neither input used (Sections~\ref{sec:dedup-cond} and~\ref{sec:dedup-blind}).
\section{Theory}
\label{sec:spec-theory}

Section~\ref{sec:spec-affine} writes the time of a step from the bytes each kernel requests and its effective bandwidth. Sections~\ref{sec:spec-cond} and~\ref{sec:spec-stair} state, in that step time, when a draft length is faster than plain decoding. They also state how the best draft length moves with acceptance. Section~\ref{sec:spec-dedup} specializes to a deduplicating engine. Section~\ref{sec:spec-err} bounds how a bandwidth error moves each prediction. Byte counts assume that the verified tokens are routed like consecutive tokens of a stationary process. Generated text that repeats its own earlier phrases violates this assumption (Section~\ref{sec:dedup-bytes}). Appendix~\ref{app:spec-proofs} gives the proofs, with the Lean theorems that check them.

\subsection{Step time}
\label{sec:spec-affine}

The verify time of a pass of $n$ tokens on kernel $p$ is
\begin{equation}
\tau_p(n)=t_0+\frac{D(n)}{\beta_D}+\frac{E_p(n)}{\beta_p(n)}+\phi_p(n),
\label{eq:tau}
\end{equation}
where $t_0\ge0$ is time outside the memory reads and $D(n)=D+\kv(n-1)$ counts the bytes read once per pass at the dense bandwidth $\beta_D$. The kernel requests the expert bytes $E_p(n)$: $\Es(n)$ for the per-slot kernel, and $\Eu(n)$ times a byte factor $f_l\ge1$ for the deduplicating kernel. The factor $f_l$ is the bytes that the deduplicating kernel reads per distinct expert of layer $l$, over $w_l$. Finally, $\phi_p(n)\ge0$ is fixed work, such as launches. The fit~\eqref{eq:reread} describes the expert part of~\eqref{eq:tau} for one layer of the per-slot kernel. Summed over layers, $c_{\mathrm{u}}Ub+c_{\mathrm{re}}(nm-U)b$ is $\Es(n)/\beta_{\mathrm{slot}}(n)$, and $t_{\mathrm{op}}$ is that layer's share of $\phi_{\mathrm{slot}}(n)$. The terms $t_0$ and $D(n)/\beta_D$ lie outside the expert operation and have no counterpart in~\eqref{eq:reread}. The bandwidths of Table~\ref{tab:beta} divide the bytes by the whole operation time, so they include $t_{\mathrm{op}}$. The predictions of Section~\ref{sec:dedup-cond} use those operation times directly. So the split between $\beta_p$ and $\phi_p$ does not enter them. A one-token pass runs the same kernel either way. So $\tau(1)$, $\beta(1)$ and $\phi(1)$ carry no kernel index. The step time is $V(k)=\tau_p(k+1)+h(k)$, with $h(k)$ the time per step of the draft model and the host. Any measured time has some $\beta_p(n)$. So~\eqref{eq:tau} says something only through where the bandwidth comes from (Section~\ref{sec:motivation}).

Take $f_l=1$ and equal fixed work. Then $t_0$ and the dense reads cancel in the saving $S(n)=\tau_{\mathrm{slot}}(n)-\tau_{\mathrm{dedup}}(n)$. Because $\Eu/\beta_{\mathrm{dedup}}=(\Es/\beta_{\mathrm{slot}})\,\bar u/r$, the saving is
\begin{equation}
S(n)=\frac{\Es(n)}{\beta_{\mathrm{slot}}(n)}\Bigl(1-\frac{\bar u(n)}{r(n)}\Bigr);
\label{eq:saving-ratio}
\end{equation}
unequal fixed work subtracts $\phi_{\mathrm{dedup}}(n)-\phi_{\mathrm{slot}}(n)$. The deduplicating kernel is faster exactly when
\begin{equation}
\bar u(n)<r(n).
\label{eq:grp-faster}
\end{equation}
The scale $\Es/\beta_{\mathrm{slot}}$ is the expert time of the per-slot kernel. One bandwidth for both kernels, $r\equiv1$, makes the deduplicating kernel faster at every $n\ge2$.

\subsection{Speedup condition}
\label{sec:spec-cond}

Take the step time $V(k)$ and throughput $T_a(k)$ of Section~\ref{sec:setting}. A draft of length $k\ge1$ adds the time $V(k)-V(0)$. It has higher throughput than plain decoding exactly when this time is less than the expected number of accepted drafts times the time of a plain step,
\begin{equation}
T_a(k)>T_a(0)\iff V(k)-V(0)<(a+a^2+\dots+a^k)\,V(0),
\label{eq:speedup-cond}
\end{equation}
which with $h=0$ reads $\tau_p(k+1)-\tau(1)<(a+\dots+a^k)\,\tau(1)$ (Eq.~\eqref{eq:app-cond-k}). The speedup is
\begin{equation}
\frac{T_a(k)}{T_a(0)}=\frac{G_a(k)\,V(0)}{V(k)}.
\label{eq:speedup}
\end{equation} At $k=1$ the condition reads $a>a_0=V(1)/V(0)-1$. So $a_0$ is the acceptance above which one draft is faster than none. The condition holds for any positive step time and assumes no common bandwidth. By~\eqref{eq:tau}, $V(k)$ carries the bandwidth $\beta_p(k+1)$ of a pass of $k+1$ tokens. Likewise, $V(0)$ carries the bandwidth $\beta(1)$ of a one-token pass. Each bandwidth is measured at its own pass size.
\label{sec:spec-speedup}

\subsection{Staircase}
\label{sec:spec-stair}

The best draft length never decreases as acceptance rises, $a\le a'\Rightarrow k^*(a)\le k^*(a')$. This holds because $G_a(k)/G_a(i)$ is non-decreasing in $a$ for $i<k$. The result does not assume a single peak of $T_a$ in $k$, so the staircase can jump by more than one length. Each step point is exact: $a_j=\max_{i<j}\min_{k\ge j}\alpha_{k,i}$, with $\alpha_{k,i}$ the least acceptance above which draft length $k$ has higher throughput than $i$ (Eqs.~\eqref{eq:app-stair-ratio} to~\eqref{eq:app-aj}). If the step time is also convex and non-decreasing in $k$ ($0\le c_1\le c_2\le\dots$), then $j$ and $j-1$ alone decide whether $k^*(a)$ reaches $j$,
\begin{equation}
k^*(a)\ge j\iff a^j\,V(j-1)>G_a(j-1)\,c_j ,
\label{eq:stair-step}
\end{equation}
and $a_1=\min(c_1/V(0),1)$ (Eqs.~\eqref{eq:app-inc-i} and~\eqref{eq:app-inc-ii}). The algebra of this subsection and the previous one uses only $V>0$. So these results hold for any positive step time, and the Lean statements are proved in that generality.

\subsection{Deduplicating engines}
\label{sec:spec-dedup}

The \emph{byte-weighted mean reuse} of the $k$ drafted tokens is
\begin{equation}
\rbar_k=\frac{1}{kW}\sum_l w_l\sum_{j=2}^{k+1}\rho_l(j),
\label{eq:rbar}
\end{equation}
and by~\eqref{eq:union-reuse} the expert bytes of a pass of $k+1$ tokens on a deduplicating kernel are
\begin{equation}
\Eu(k+1)=mW+mWk(1-\rbar_k),
\label{eq:Eu-rbar}
\end{equation}
a count that involves the routing alone. Take a deduplicating engine with $f_l=1$ and $h=0$. Write $B_k=\beta_{\mathrm{dedup}}(k+1)$, $\Delta D_k=D(k+1)-D(1)$ and $\Delta\phi_k=\phi_{\mathrm{dedup}}(k+1)-\phi(1)$. By~\eqref{eq:tau}, a draft of length $k$ adds the time
\begin{equation}
V(k)-V(0)=\frac{\Delta D_k}{\beta_D}+\Delta\phi_k+\frac{\Eu(k+1)}{B_k}-\frac{mW}{\beta(1)} ,
\label{eq:dV-time}
\end{equation}
and by~\eqref{eq:speedup-cond} it is faster than plain decoding exactly when $\rbar_k>\rho^*_k(a)$, with
\begin{equation}
\rho^*_k(a)=1-\frac{B_k}{mWk}\Bigl[(G_a(k)-1)\,\tau(1)-\frac{\Delta D_k}{\beta_D}-\Delta\phi_k\Bigr]-\frac1k\Bigl(\frac{B_k}{\beta(1)}-1\Bigr)
\label{eq:rho-star}
\end{equation}
(Eq.~\eqref{eq:app-time-iff}). The threshold falls as $a$ grows, since $G_a(k)$ rises with $a$ and $\tau(1),B_k>0$ (Eq.~\eqref{eq:app-time-move}). The bracket is the time that a draft may add outside the expert reads. The bandwidth $B_k$ converts this time into union bytes. The last term asks for more reuse when the deduplicating kernel runs slower on $k+1$ tokens than on one. Substituting~\eqref{eq:dV-time} into the speedup~\eqref{eq:speedup} gives
\begin{equation}
\frac{T_a(k)}{T_a(0)}=\frac{G_a(k)\,V(0)}{V(0)+\Delta D_k/\beta_D+\Delta\phi_k+\Eu(k+1)/B_k-mW/\beta(1)},
\label{eq:speedup-dedup}
\end{equation}
with the bytes $\Eu(k+1)$ fixed by the routing traces before any deduplicating kernel runs, and the bandwidths $B_k$ and $\beta(1)$ measured. With one bandwidth for every byte and $t_0=\phi=0$, \eqref{eq:rho-star} reduces to the byte form
\begin{equation}
\rho^*_k(a)=1-\frac{(G_a(k)-1)\Vu(0)-\bigl(\kv(k)-\kv(0)\bigr)}{mWk},
\label{eq:rho-star-bytes}
\end{equation}
with $\Vu(0)=D+mW+\kv(0)$ the bytes of a plain step (Eq.~\eqref{eq:app-time-bytes}). This is the threshold that a pure byte count would give. When $a>0$, the threshold~\eqref{eq:rho-star} also falls as $D$ grows, provided the bandwidths and the fixed work do not change with $D$ (Eq.~\eqref{eq:app-time-move}). More reuse lowers every $\Eu(k+1)$. So a speedup persists under any routing with at least as much reuse. This holds provided that the bandwidth and the fixed work of the deduplicating kernel at fixed $n$ do not depend on which experts repeat. Under the same proviso, routing enters only through $(\rbar_1,\dots,\rbar_{\kmax})$. Also, no single rule ``$\rbar_k>c$'' decides for every routing (Appendix~\ref{app:spec-dedup-sec}).

\subsection{Error bounds}
\label{sec:spec-err}

Let $\hat\beta=\beta(1+e)$ for each bandwidth in~\eqref{eq:tau}, with $|e|\le\varepsilon<1$ and the other terms exact. Here $\varepsilon$ is the bandwidth error of Section~\ref{sec:motivation}. Write $M_p(n)=D(n)/\beta_D+E_p(n)/\beta_p(n)$ for the memory part of $\tau_p(n)$ and $M^{\mathrm{e}}_p(n)=E_p(n)/\beta_p(n)$ for its expert part. A hat marks a quantity computed from the estimated bandwidths: $\hat M$, $\hat\tau_p$, $\hat S$, $\hat V$, $\hat T_a=G_a/\hat V$, and $\hat k^*$, the draft length that maximizes $\hat T_a$. For the best draft length and the best speedup, $\delta$ bounds the relative error of the full predicted step time, $(1-\delta)V(k)\le\hat V(k)\le(1+\delta)V(k)$. With bandwidth errors alone, $\delta\le\varepsilon/(1-\varepsilon)$. Then a 10\% bandwidth error moves the predicted speedup by at most 25\%. Table~\ref{tab:bounds} states the five bounds, each proved in Lean (Appendix~\ref{app:err-proofs}). Relative to the saving, the saving bound~\eqref{eq:e2} is $\varepsilon/(1-\varepsilon)$ times the sum of two expert times over their difference. This factor grows without limit near the crossover. A verify time carries the factor $M_p/\tau_p\le1$, and a common factor on every $V(k)$ leaves $\hat k^*=k^*$. So the saving is the prediction most sensitive to a bandwidth error, and the best draft length is the least sensitive.

\begin{table}[t]\centering\small

\caption{Error bounds for a relative bandwidth error $|e|\le\varepsilon$ (first three rows) or a relative step-time error $\delta$ (last two).}
\label{tab:bounds}
\renewcommand{\arraystretch}{1.35}
\begin{tabular}{lp{0.55\textwidth}}
\toprule
prediction & bound \\
\midrule
verify time & $|\hat\tau_p(n)-\tau_p(n)|\le\frac{\varepsilon}{1-\varepsilon}M_p(n)\le\frac{\varepsilon}{1-\varepsilon}\tau_p(n)$ \hfill\refstepcounter{equation}(\theequation)\label{eq:e1}\\
saving & $|\hat S(n)-S(n)|\le\frac{\varepsilon}{1-\varepsilon}\bigl(M^{\mathrm{e}}_{\mathrm{slot}}(n)+M^{\mathrm{e}}_{\mathrm{dedup}}(n)\bigr)$ \hfill\refstepcounter{equation}(\theequation)\label{eq:e2}\\
faster kernel & sides of~\eqref{eq:grp-faster} agree if $|\ln r(n)-\ln\bar u(n)|>\ln\frac{1+\varepsilon}{1-\varepsilon}$ \hfill\refstepcounter{equation}(\theequation)\label{eq:e3}\\
best draft length & $T_a(\hat k^*)\ge\frac{1-\delta}{1+\delta}T_a(k^*)$: at most $\frac{2\delta}{1+\delta}$ of the throughput lost; $\hat k^*=k^*$ if $\frac{1+\delta}{1-\delta}T_a(i)<T_a(k^*)$ for every $i\ne k^*$ \hfill\refstepcounter{equation}(\theequation)\label{eq:e4}\\
best speedup & $\frac{1-\delta}{1+\delta}\le\frac{\hat T_a(\hat k^*)/\hat T_a(0)}{T_a(k^*)/T_a(0)}\le\frac{1+\delta}{1-\delta}$ \hfill\refstepcounter{equation}(\theequation)\label{eq:e5}\\
\bottomrule
\end{tabular}
\end{table}
\section{Experiments}
\label{sec:dedup}

\label{sec:dedup-forks}\label{sec:dedup-reg}Each engine gets one switch that sends small verification passes to its existing deduplicating expert kernel. With \emph{deduplication off}, the engine keeps its default kernel choice: the per-slot kernel at $n\le9$, or at $n\le8$ on CUDA, which runs the deduplicating kernel at $n=9$. With \emph{deduplication on}, it runs the deduplicating kernel from $n=2$ (Appendix~\ref{app:dedup-evidence}, Table~\ref{tab:forks}). Appendix~\ref{app:dedup-evidence} gives the change to each engine and the check that both settings produce the same output.

Each configuration has a context depth $d\in\{512,4096\}$, the number of prompt tokens before decoding starts. OLMoE, Qwen3 and gpt-oss give twelve configurations. In each configuration we measure two quantities on the same six prompts. These prompts are articles of the WikiText-103 validation set truncated to $d$ tokens, disjoint from the test-set documents that give the routing traces. The first quantity is the verify time with deduplication off and on at $n=1,\dots,9$. The second is the throughput of speculative decoding at draft lengths $k=0,\dots,8$ and 11 acceptance values from 0.2 to 0.98. It uses the \emph{oracle draft}, which proposes the target's own greedy token and corrupts each draft with probability $1-a$. So acceptance is geometric at rate $a$ (Appendix~\ref{sec:spec-tests}). The \emph{prediction} computes the time of each layer's expert operation from its bytes. We made the fit~\eqref{eq:reread} and its deduplicating counterpart on the passes with chosen ids of Section~\ref{sec:motivation}. We then evaluated each fit at the union bytes of every window of recorded routing from the WikiText-103 traces. So the prediction uses nothing from the whole-model runs. Then~\eqref{eq:cond} below sums the layers into a whole-model verify time. Table~\ref{tab:compare} also gives the prediction with the recorded routing of the text those runs verified. The same sum with measured operation times is the \emph{measured-time version}. It skips the fit and uses each layer's measured median operation time on each window of recorded routing. It serves where a model has no fit. The gap between the measured-time version and the prediction is the fits' error on recorded routing, which $\varepsilon$ bounds. On OLMoE, Qwen3 and gpt-oss, the prediction and the measured-time version were computed after the whole-model data existed. As first computed, the prediction took the one-token time $\bar t_l(1)$ from the deduplicating kernel's fit. It then placed the crossover wrongly in ten of twelve configurations. We corrected it to the per-slot fit, because the engine runs the per-slot kernel at $n=1$ with deduplication off and on. This was the only change made after a comparison with the measurements. DeepSeek-V2-Lite had no such fit. On it, we wrote down the measured-time version and its \emph{scoring rules}, the tolerances it had to meet, before any whole-model speculative run (Section~\ref{sec:dedup-blind}).

\label{sec:dedup-beta}\label{sec:dedup-cond}The prediction and the measured-time version both take the whole-model verify time from the expert operation,
\begin{equation}
\hat\tau_p(n)=\tau^{\mathrm{meas}}(1)+\sum_{l=1}^{L}\bigl(\bar t_{p,l}(n)-\bar t_l(1)\bigr),
\label{eq:cond}
\end{equation}
where $\tau^{\mathrm{meas}}(1)$ is the measured whole-model time of a one-token pass. The term $\bar t_{p,l}(n)$ is layer $l$'s operation time with kernel $p$, averaged over the windows of recorded routing. On the Apple engines, $n=9$ is extrapolated. On OLMoE, Qwen3 and gpt-oss, the measured-time version uses the recorded routing of the text the whole-model runs verified. The term $\bar t_l(1)$ is the same layer's one-token time, which the per-slot kernel runs with deduplication off and on. The predicted expert time is $\hat M^{\mathrm{e}}_p(n)=\sum_l\bar t_{p,l}(n)$. The saving, the crossover, the best draft length $\hat k^*$ and the speedup follow, with $\hat V(k)=\hat\tau_{\mathrm{dedup}}(k+1)+h(k)$ and $h(k)$ measured. The \emph{calibrated version} is the measured-time version with one change. It takes the growth with $n$ of the time outside the expert operation from the measured verify time with deduplication off. Besides caching and tokens per expert, fixed work separates the two kernels. With deduplication on, CUDA makes 160 kernel launches per OLMoE pass against 64, a difference of about 0.1 ms.

We compute the bandwidth error $\varepsilon$ from the fits of Section~\ref{sec:motivation}. Here $\varepsilon_p$ is the median, over layers and models, of each fit's median relative error on the windows of recorded routing from the WikiText-103 traces. The engine's $\varepsilon$ is the larger of its two kernels' values: 0.054 on llama.cpp CUDA, 0.031 on llama.cpp Metal and 0.148 on MLX-LM. The measurement takes under an hour per model. For DeepSeek-V2-Lite it took 22 minutes on CUDA and 48 on Metal, against 3.8 and about 8 hours of whole-model speculative runs.

\subsection{Bytes}
\label{sec:dedup-bytes}

\begin{figure}[t]
\centering

\includegraphics[width=0.85\linewidth]{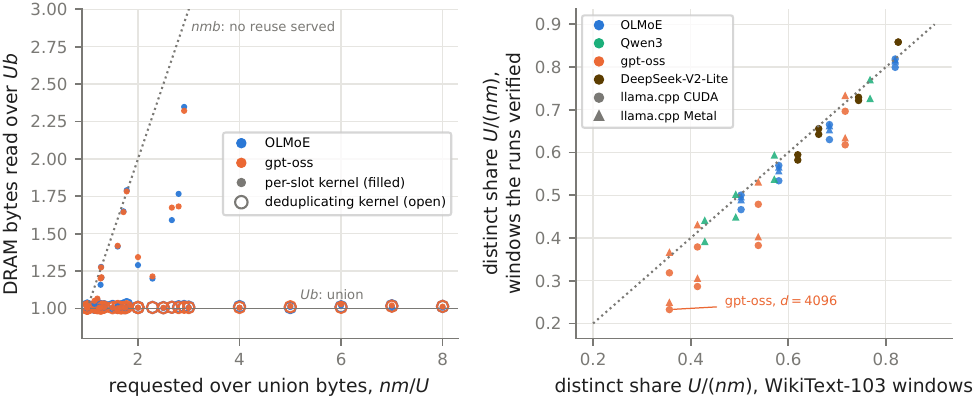}
\caption{Bytes. Left: DRAM bytes read by the CUDA expert kernels over the union bytes $Ub$, against $nm/U$ (Nsight Compute, chosen ids, one layer of OLMoE and gpt-oss). Filled: per-slot kernel; open: deduplicating kernel; color: model. The deduplicating kernel reads $Ub$ (horizontal line). The per-slot kernel reads between $Ub$ and its requested $nmb$ (dotted). Right: the share of a pass's $nm$ slots that name distinct experts, $U/(nm)$, at $n=2,4,6,8$. It is shown on the windows the whole-model runs verified, against WikiText-103 windows of the same model. Both axes lie between $1/n$ (every token repeats the first token's experts) and 1 (no expert repeats). A point above the dotted line means the verified text repeated experts less often than WikiText-103. Color: model; shape: the engine of the runs (circle: llama.cpp CUDA, triangle: llama.cpp Metal). The labeled gpt-oss points at $d=4096$ below the line are its repeating continuations.}
\label{fig:bytes}
\end{figure}

We profiled the CUDA expert kernels with Nsight Compute \citep{nsightcompute} on passes with chosen ids at $n=2,\dots,8$, on one layer of OLMoE and gpt-oss. In all 70 profiles, the deduplicating kernel reads the union bytes $Ub$ from DRAM within 0.2 to 2.5\%. The per-slot kernel requests $nmb$ and reads between $Ub$ and $2.35\,Ub$ (Figure~\ref{fig:bytes}). Here the L2 cache serves part of the repeated slots, an effect that belongs to the per-slot kernel's effective bandwidth. The union is as good as the routing it is given. On the windows the whole-model runs verified, the share of distinct slots $U/(nm)$ agrees with that of WikiText-103 windows within 0.06. This holds at every $n$ on every model except gpt-oss at $d=4096$. There the greedy continuations repeat, and the share lies 0.08 to 0.16 below: 81\% of their 4-grams recur, against 48\% at $d=512$.

\subsection{Saving and crossover}
\label{sec:dedup-cross}

The prediction's largest saving error is 0.23 to 0.52 ms on CUDA and 0.43 to 3.7 ms on the Apple engines (Table~\ref{tab:compare}). The exception is Metal gpt-oss at $d=4096$, with 9.7 ms, whose verified text repeats and routes unlike WikiText-103 (Figure~\ref{fig:bytes}). Its crossover is within one of the measured one in 10 of 12 configurations. It equals the measured crossover in all six other Apple configurations and is one pass size early in three of the four CUDA configurations. It misses both Metal gpt-oss configurations. With the recorded routing of the verified text, the saving error is 0.25 to 0.51 ms on CUDA and 0.42 to 3.8 ms on the Apple engines. The crossover is then within one in 11 of 12 configurations. The measured-time version gives saving errors of 0.17 to 0.20 ms on CUDA and 0.8 to 3.9 ms on the Apple engines. Its crossover is within one in 11 of 12 configurations and equal in all four CUDA configurations. The remaining miss is Metal gpt-oss at $d=512$. Its measured saving at $n=8$ is $+0.2$ ms, with an interval containing 0, and there the margin of~\eqref{eq:e3} fails.

\begin{table}[t]\centering\small

\caption{The prediction with WikiText-103 routing (wiki) and with the routing of the verified text (text), and the measured-time version (op.\ time). All three are compared with the same measurements, one row per configuration. Saving error: largest $|\hat S(n)-S(n)|$ over $n=2,\dots,9$ ($n\le8$ on CUDA), in ms. Crossover as defined in Section~\ref{sec:motivation} (meas.: measured; $-$: none). $k^*$: acceptance values, of 11, at which the predicted best draft length is within one of the measured one. Metal and CUDA are llama.cpp.}
\label{tab:compare}
\renewcommand{\arraystretch}{1.1}\setlength{\tabcolsep}{4pt}
\begin{tabular}{lll ccc cccc ccc}
\toprule
& & & \multicolumn{3}{c}{saving error (ms)} & \multicolumn{4}{c}{crossover} & \multicolumn{3}{c}{$k^*$ within one} \\
\cmidrule(lr){4-6}\cmidrule(lr){7-10}\cmidrule(lr){11-13}
engine & model & $d$ & wiki & text & op.\ time & meas. & wiki & text & op.\ time & wiki & text & op.\ time \\
\midrule
Metal & OLMoE & 512 & 0.43 & 0.42 & 0.77 & 7 & 7 & 7 & 8 & 9 & 9 & 9 \\
Metal & OLMoE & 4096 & 0.98 & 0.91 & 2.28 & 7 & 7 & 7 & 8 & 8 & 8 & 8 \\
Metal & Qwen3 & 512 & 1.51 & 0.92 & 2.24 & $-$ & $-$ & $-$ & $-$ & 7 & 8 & 7 \\
Metal & Qwen3 & 4096 & 3.70 & 1.83 & 1.25 & $-$ & $-$ & $-$ & $-$ & 8 & 8 & 8 \\
Metal & gpt-oss & 512 & 2.23 & 3.77 & 3.89 & 8 & $-$ & $-$ & $-$ & 10 & 9 & 10 \\
Metal & gpt-oss & 4096 & 9.72 & 3.46 & 3.89 & 6 & $-$ & 7 & 7 & 10 & 9 & 10 \\
MLX-LM & OLMoE & 512 & 2.60 & 2.86 & 3.29 & $-$ & $-$ & $-$ & $-$ & 8 & 7 & 8 \\
MLX-LM & OLMoE & 4096 & 2.51 & 2.64 & 2.37 & $-$ & $-$ & $-$ & $-$ & 7 & 7 & 9 \\
CUDA & OLMoE & 512 & 0.23 & 0.25 & 0.20 & 5 & 4 & 4 & 5 & 11 & 11 & 11 \\
CUDA & OLMoE & 4096 & 0.26 & 0.26 & 0.17 & 5 & 4 & 4 & 5 & 11 & 11 & 11 \\
CUDA & gpt-oss & 512 & 0.35 & 0.49 & 0.18 & 6 & 5 & 5 & 6 & 10 & 10 & 11 \\
CUDA & gpt-oss & 4096 & 0.52 & 0.51 & 0.19 & 5 & 5 & 4 & 5 & 10 & 11 & 11 \\
\bottomrule
\end{tabular}
\end{table}

\begin{figure}[tp]
\centering

\includegraphics[width=0.8\linewidth]{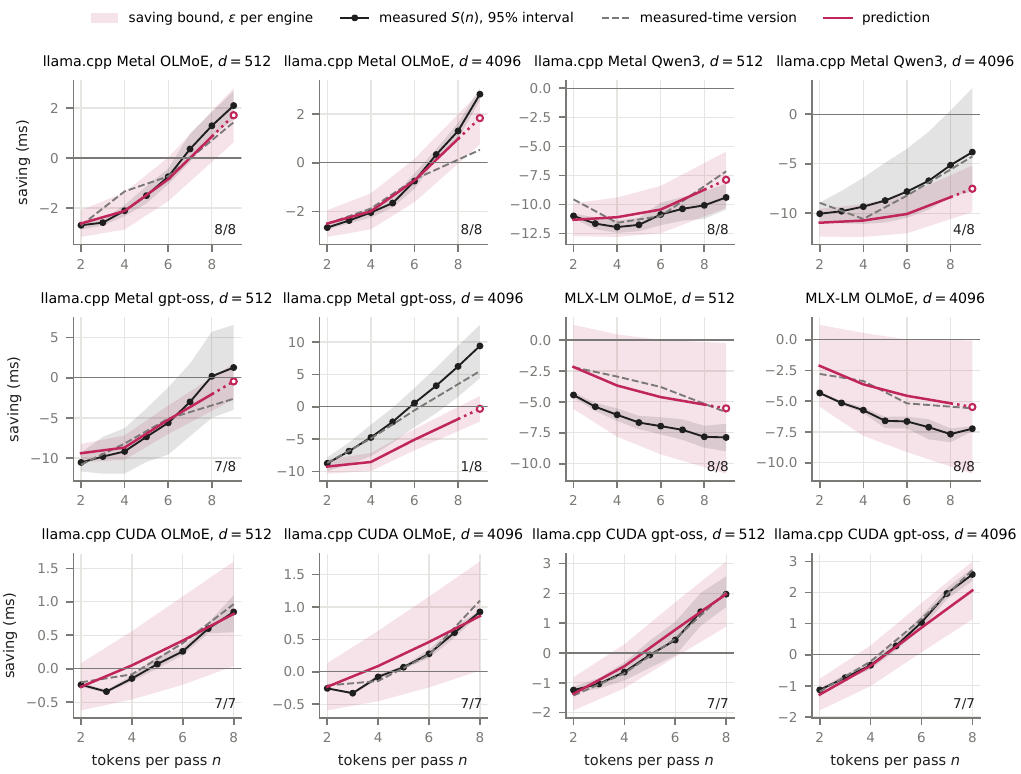}
\caption{Saving $S(n)$. Black and grey: measured on the six validation prompts, with its 95\% bootstrap interval. Red: the prediction, dotted to an open circle where $n=9$ is extrapolated. Shaded red: the band of the saving bound~\eqref{eq:e2} for the engine's $\varepsilon$. Grey dashed: the measured-time version. The number in each panel counts the measured points inside the band. Positive: the deduplicating kernel is faster. CUDA is shown to $n=8$, since it runs the deduplicating kernel at $n=9$ with deduplication off and on.}
\label{fig:saving}
\end{figure}

Figure~\ref{fig:saving} shows the saving itself at every $n$. The saving is the difference of two expert times of 2.8 to 40.0 ms each. So the saving bound~\eqref{eq:e2} amplifies a bandwidth error 3.2 to 3.8 times on MLX-LM, 6 to 220 times on CUDA and 3.2 to 340 times on Metal. The amplification is largest at the crossover (Appendix~\ref{app:dedup-evidence}, Table~\ref{tab:perconfig}). With each engine's $\varepsilon$, the measured saving lies inside the band around the prediction at all 28 CUDA points, all 16 MLX-LM points and 36 of 48 Metal points. Consider the bandwidth error at which the band just covers a configuration's largest saving error. It is 0.022 to 0.033 on CUDA and 0.107 to 0.108 on MLX-LM, below the engines' $\varepsilon$. On Metal it is 0.014 to 0.126. That is above Metal's 0.031 in three of six configurations, and most above it on gpt-oss at $d=4096$. At $n=1$, an Apple engine runs the same kernel with deduplication off and on. There the engine with deduplication on measures slower in 6 of 8 configurations. So part of every Apple saving is a run-to-run offset.

\subsection{Best draft length and speedup}
\label{sec:dedup-kstar}

On CUDA, the predicted $\hat k^*$ is within one of the measured best at 10 or 11 of 11 acceptance values in every configuration. The measured-time version places it within one at all 11 (Table~\ref{tab:compare}). Drafting the measured-time version's $\hat k^*$ loses at most 1.6 to 7.7\% of the best measured throughput. On the Apple engines both place $\hat k^*$ within one at 7 to 10 of 11 values. An error common to every step time leaves $\hat k^*$ unchanged (Section~\ref{sec:spec-err}). On the Apple engines, an error that grows with $n$ moves it. Neither the prediction nor the measured-time version contains the growth with $n$ of attention and the dense parts. This growth cancels in the saving and stays in the verify time. With deduplication on, the prediction's verify time is within 3 to 21\% on CUDA, 6 to 25\% on Metal and 38 to 41\% on MLX-LM. The measured-time version's is within 4 to 16\% on CUDA and 15 to 31\% on the Apple engines. The calibrated version's is within 1 to 6\% in 8 of 12 configurations (Appendix~\ref{app:dedup-evidence}, Table~\ref{tab:perconfig}).

\begin{figure}[t]
\centering

\includegraphics[width=0.9\linewidth]{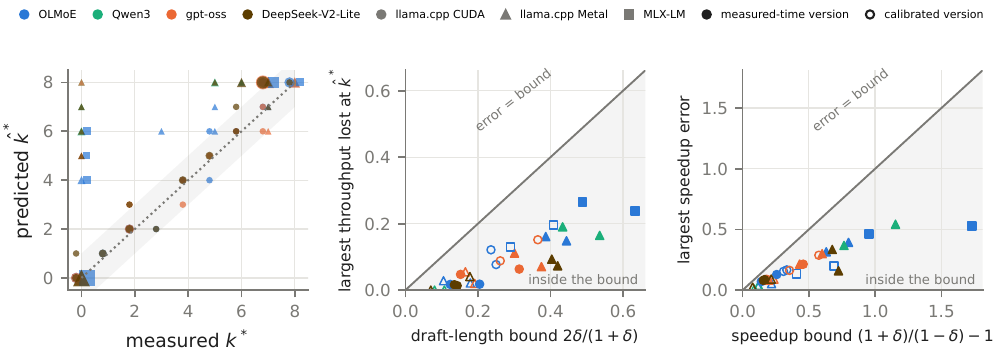}
\caption{Best draft length and the bounds~\eqref{eq:e4} and~\eqref{eq:e5}. Each configuration has one point under the measured-time version (filled) and one under the calibrated version (open), with $\delta$ as in Section~\ref{sec:dedup-kstar}. Color: model; shape: engine (circle: llama.cpp CUDA, triangle: llama.cpp Metal, square: MLX-LM). Left: predicted against measured $k^*$, measured-time version; area grows with the number of acceptance values; dotted: exact prediction; shaded: within one. Middle: largest share of throughput lost by drafting $\hat k^*$, against its bound~\eqref{eq:e4}. Right: largest relative error of the predicted best speedup, against its bound~\eqref{eq:e5}. In the middle and right panels the solid line is where the error equals its bound. The theory requires every point to lie in the shaded region on or below it, and every point does.}
\label{fig:bounds}
\end{figure}

We set $\delta$ of~\eqref{eq:e4} to the largest relative error of each configuration's step time, under the measured-time version and under the calibrated version. This gives 24 pairs of a configuration and a version, plus 6 DeepSeek-V2-Lite pairs. At all 330 acceptance values of these pairs, the throughput lost lies inside its bound~\eqref{eq:e4} and the best speedup inside its bound~\eqref{eq:e5} (Figure~\ref{fig:bounds}). Measured speculative runs on CUDA differ from the speedup computed from verify times by a median factor of 0.93 to 1.04 per configuration. Part of the reason is that the oracle draft emits 2 to 5\% fewer tokens per step than $G_a(k)$ (Appendix~\ref{app:dedup-evidence}).

\subsection{DeepSeek-V2-Lite}
\label{sec:dedup-blind}

We use DeepSeek-V2-Lite-Chat in the 4-bit llama.cpp format Q4\_K\_M \citep{deepseekv2}. It has 64 routed experts (6 per token), two shared experts that every token uses, multi-head latent attention and a dense first layer. The other three models have none of these. On llama.cpp CUDA and Metal, the measured-time version follows from~\eqref{eq:cond}. Its inputs are runs of the expert operation of all 26 MoE layers on WikiText-103 windows, and three whole-model numbers at $k=0$. We wrote the measured-time version, six scoring rules and the scoring script before the first speculative run on each engine. The script then scored the runs unchanged.

\begin{table}[t]\centering\small

\caption{DeepSeek-V2-Lite-Chat: the measured-time version and the calibrated version, written before its speculative runs. Crossover: predicted / measured ($-$: none), over $n\le8$ on CUDA. Sign: values of $n$ with the measured sign of $S(n)$, of those whose interval excludes 0. Saving error in ms. $k^*$: acceptance values, of 11, with $\hat k^*$ within one of the measured best. Loss: largest share of the best throughput lost at $\hat k^*$. $\tau$ error: largest relative error of the predicted verify time, over both kernels and $n=2,\dots,9$. $\delta$ as in Section~\ref{sec:dedup-kstar}. The best speedup lies inside its bound~\eqref{eq:e5} in every row.}
\label{tab:blind}
\renewcommand{\arraystretch}{1.12}\setlength{\tabcolsep}{4pt}
\begin{tabular}{lllcccccc}
\toprule
engine & version & $d$ & crossover & sign & saving error & $k^*$ within 1 & loss & $\tau$ error / $\delta$ \\
\midrule
llama.cpp CUDA & measured-time & 512 & $-$ / $-$ & 7 of 7 & 0.23 & 11 & 2.9\% & 9.9\% / 0.072 \\
 & measured-time & 4096 & $-$ / $-$ & 7 of 7 & 0.28 & 11 & 4.1\% & 10.1\% / 0.077 \\
llama.cpp Metal & measured-time & 512 & $-$ / $-$ & 8 of 8 & 1.6 & 9 & 20.1\% & 34.0\% / 0.27 \\
 & measured-time & 4096 & $-$ / $-$ & 8 of 8 & 4.4 & 6 & 22.5\% & 28.4\% / 0.25 \\
 & calibrated & 512 & $-$ / $-$ & 8 of 8 & 1.5 & 11 & 6.4\% & 4.3\% / 0.036 \\
 & calibrated & 4096 & $-$ / $-$ & 8 of 8 & 4.4 & 9 & 8.3\% & 10.5\% / 0.098 \\
\bottomrule
\end{tabular}
\end{table}

On CUDA, the deduplicating kernel is predicted and measured slower at every $n$ from 2 to 8. The saving lies inside its band at all 14 points, and the best draft length is within one at all 22 acceptance values. The verify time is within 10.1\% (Table~\ref{tab:blind}). On Metal, the sign of the saving is right at every $n$, but the verify time is predicted up to 34\% too low. The reason is the growth with $n$ of attention, the shared experts and the dense layer. This growth is large on this model and lies outside the expert operation. The error grows with $n$, so drafting $\hat k^*$ loses up to 22.5\% of the best throughput. The calibrated version was written down beforehand as a second prediction. It brings the verify time within 4.3\% and 10.5\%, and the best draft length within one at 20 of 22 values. The Metal saving requires a bandwidth error of up to 0.075, 2.4 times Metal's $\varepsilon$ of 0.031. Of the six scoring rules (Appendix~\ref{app:dedup-evidence}, Table~\ref{tab:blind-criteria}), the CUDA measured-time version meets all six at both depths. The Metal measured-time version meets the crossover, sign and speedup rules. It fails the rules on the best draft length, the throughput lost and the verify time. On CUDA, the measured throughput speedup lies 1 to 21\% below the predicted one. This is the gap of the oracle draft (Section~\ref{sec:dedup-kstar}), which~\eqref{eq:e5} does not cover.

\begin{figure}[t]
\centering

\includegraphics[width=\linewidth]{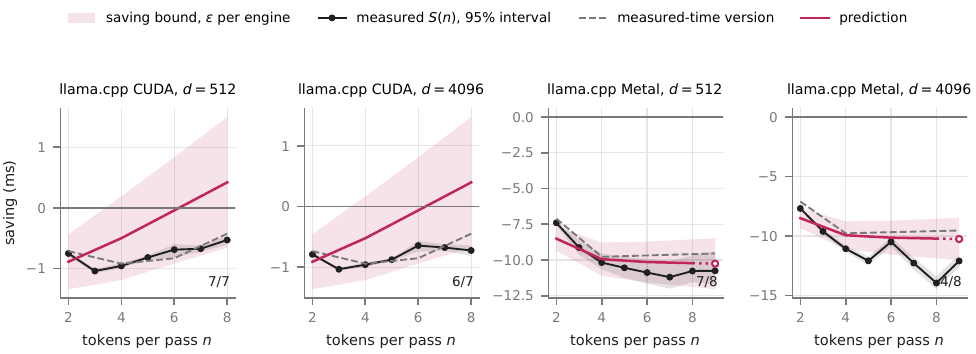}
\caption{Saving $S(n)$ on DeepSeek-V2-Lite, as in Figure~\ref{fig:saving}. Black and grey: measured, with its 95\% interval. Red: the prediction from fits run after the whole-model runs, with the band of the saving bound~\eqref{eq:e2} for the engine's $\varepsilon$. Grey dashed: the measured-time version, written before its speculative runs. The number in each panel counts the measured points inside the band.}
\label{fig:deepseek}
\end{figure}

The prediction needs fits on chosen ids. We ran them on DeepSeek-V2-Lite afterwards, on layers 1 and 13, with the protocol of the other models. We then applied the chain unchanged to the same WikiText-103 windows. The whole-model results were known by then, and we adjusted nothing in the chain. The fits' errors on recorded routing are 0.021 to 0.049 on Metal and 0.027 to 0.129 on CUDA. On Metal, the prediction matches or improves on the measured-time version at both depths. Its saving error is 1.1 and 3.7 ms, against 1.6 and 4.4. No crossover is predicted or measured. Both place the best draft length within one at 9 and 6 of 11 values. The prediction's verify time is within 31.6\% and 26.4\%, against 34.0\% and 28.4\%, over both kernels and $n=2,\dots,9$ as in Table~\ref{tab:blind}. On CUDA, the prediction keeps the verify time within 8.0\% and 8.3\%, and the best draft length within one at all 22 values. But the per-slot fit predicts the windows of recorded routing 5 to 13\% slower than they run. That adds about 0.9 ms of per-slot time at $n=8$, against a measured saving of about $-0.5$ ms. So the prediction places a crossover at $n=7$ where none is measured. Its saving errors are 0.95 and 1.1 ms, against 0.23 and 0.28 ms for the measured-time version (Figure~\ref{fig:deepseek}). With the engines' $\varepsilon$ from Section~\ref{sec:dedup-beta}, the measured saving lies inside the band around the prediction at 13 of 14 CUDA points and 11 of 16 Metal points.

\FloatBarrier
\section{Limits}
\label{sec:discussion}

The theory assumes geometric acceptance, verified tokens routed like consecutive tokens of a stationary process, and no cost for the draft model. The theory does not predict the bandwidth of any kernel. We measure each kernel's bandwidth on each model's expert shapes. Bandwidth fitted on the other models of an engine misses a held-out model by 9 to 206\% (Appendix~\ref{app:dedup-evidence}). The prediction on OLMoE, Qwen3 and gpt-oss was computed after their whole-model data existed. On DeepSeek-V2-Lite, the prediction used fits run after the whole-model runs. Only the measured-time and calibrated versions on DeepSeek-V2-Lite were written before its speculative runs. On OLMoE, Qwen3 and gpt-oss, the measured-time version uses the recorded routing of the verified continuations, which approximates that of verify windows containing drafts. On Metal, the times of the expert operation alone miss part of each kernel's growth inside the whole model. The sign of this miss depends on the model. From $n=7$, the per-slot kernel exceeds its growth on the expert operation alone on gpt-oss and OLMoE. On DeepSeek-V2-Lite the deduplicating kernel does, by 1.1 to 4.4 ms. The Metal runs of DeepSeek-V2-Lite at $d=512$ show a recurring slow GPU state with no thermal warning logged. The scoring rules include these runs. The measurements use one machine per platform, at batch one, with four models of 7B to 30B total parameters. DRAM byte counts exist only on CUDA.

\section{Conclusion}
\label{sec:conclusion}

A speedup is a ratio of step times, and a step's time is its bytes over the effective bandwidth of the kernel that reads them. In that step time, the theory states exactly when speculation is faster, at which draft length and by how much. These results are proved in Lean. On a deduplicating engine the bytes follow from routing, and the CUDA deduplicating kernel reads them within 2.5\%. The per-slot kernel has the higher effective bandwidth on every engine and model measured. So the deduplicating kernel is faster only where $\bar u(n)<r(n)$. Its saving is the per-slot expert time times $1-\bar u(n)/r(n)$~\eqref{eq:saving-ratio}. The bandwidth's error carries through the proved bounds as stated on CUDA and MLX-LM. It moves the best draft length and the speedup little on CUDA, and the saving amplifies it. On DeepSeek-V2-Lite, operation times measured before any of its speculative runs give the CUDA best draft length within one at every acceptance value. The prediction from fits run afterwards misses its CUDA crossover.

The results change how a cost model for MoE speculation should charge expert reads. Models with one bandwidth for every byte, such as S2-MoE \citep{arXiv:2608.15018} and SpecFit \citep{arXiv:2610.15417}, imply that deduplication with equal fixed work is faster wherever $\bar u(n)<1$. That is every pass size with reuse. On every engine and model that we measured, the deduplicating kernel is slower at two tokens (Section~\ref{sec:motivation}). A cost model must therefore compare the byte share $\bar u(n)$ with the bandwidth ratio $r(n)$. The bytes alone do not decide the sign of the saving.

For engine developers, the switch between the two kernels can follow from measurement. Engines currently tune it by hand \citep{llamacpp30047}. Routing traces give $\bar u(n)$, and the benchmark of the expert operation gives $r(n)$ in under an hour per model. Together with the fixed work, they give the predicted crossover, and the deduplicating kernel should run from there. The ratio $r(n)$ is also the lever for kernel work. With equal fixed work, a deduplicating kernel at $r(n)=1$ is faster at every pass size with reuse. A higher bandwidth at small $n$ can move the crossover to fewer tokens.

Four extensions follow from the limits above. First, the bandwidth needs a model from kernel structure, since a bandwidth fitted on other models misses a held-out model. Second, a real draft model and acceptance that is not geometric would test the staircase under the conditions of deployment. Third, batches larger than one change both the routing union and the kernels' bandwidths. Fourth, hardware with lower bandwidth or weaker compute, such as edge devices, would test whether the step time stays limited by memory reads.

\section*{Code and data availability}
The modified engines, the benchmark of the expert operation, the chosen ids, the routing traces, the timing files and the Lean formalization are available from the author on request.

\section*{Use of AI}
AI agents directed by the author produced the code that ran the experiments, generated the figures and tables and consolidated the statistics. They also produced the Lean formalization, the initial literature review and the initial draft of the manuscript. The author chose the research direction, designed the experiments, reviewed the results and findings, and is responsible for the paper.

\appendix
\let\appsection\section\renewcommand{\section}{\clearpage\suppressfloats[t]\appsection}
\counterwithin*{equation}{section}\renewcommand{\theequation}{\thesection.\arabic{equation}}
\section{Notations}
\label{app:notation}
{\renewcommand{\theHtable}{notation}%
\begin{longtable}{p{0.24\textwidth}p{0.68\textwidth}}
\toprule
Symbol & Meaning\\
\midrule
\endfirsthead
\toprule
Symbol & Meaning\\
\midrule
\endhead
$a$ & acceptance rate (geometric acceptance)\\
$a_0$ & $V(1)/V(0)-1$, the acceptance above which one draft is faster than none\\
$a_j$ & $j$-th step point\\
$b$ & bytes of one expert of the timed layer, $b=w_l$\\
$B_k$ & $\beta_{\mathrm{dedup}}(k+1)$\\
$c_i$ & $V(i)-V(i-1)$, the time the $i$-th draft adds\\
$c_{\mathrm{u}}$, $c_{\mathrm{re}}$ & per-slot fit~\eqref{eq:reread}: time per byte of a first read and of a repeated slot\\
$c_{\mathrm{g}}$, $c_{\mathrm{t}}$ & deduplicating fit~\eqref{eq:app-dedup-fit}: time per union byte and per token\\
$d$ & context depth in tokens\\
$D$ & bytes read once per forward pass whatever the routing\\
$D(n)$ & bytes read once per pass of $n$ tokens, $D+\kv(n-1)$\\
$e$ & relative bandwidth error (Section~\ref{sec:spec-err})\\
$\E$ & expectation\\
$E_p(n)$ & expert bytes kernel $p$ requests on a pass of $n$ tokens\\
$\Es(n)$, $\Eu(n)$ & expected expert bytes of a pass of $n$ tokens with the per-slot and the deduplicating kernel; $\Eu$ are the union bytes\\
$f_l$ & bytes the deduplicating kernel reads per distinct expert of layer $l$, over $w_l$\\
$G_a(k)$ & $(1-a^{k+1})/(1-a)$, expected tokens emitted per step\\
$h(k)$ & draft model's and host's time per step\\
$h_j(a)$ & $a^j/G_a(j-1)$ (Appendix~\ref{app:spec-proofs})\\
$k$, $\kmax$ & draft length; its maximum\\
$k^*(a)$ & best draft length, the smallest maximizer of $T_a$\\
$\kv(k)$ & key-value cache bytes read by a step verifying $k+1$ tokens\\
$L$, $l$ & number of MoE layers; layer index\\
$m$ & experts selected per token\\
$M_p(n)$ & memory part of $\tau_p(n)$\\
$M^{\mathrm{e}}_p(n)$ & expert part of $\tau_p(n)$, $E_p(n)/\beta_p(n)$\\
$n$ & number of consecutive tokens, or tokens verified in one pass\\
$N$ & experts per layer\\
$p$ & expert kernel of a pass: slot (per-slot) or dedup (deduplicating)\\
$P_k(x)$ & $1+x+\dots+x^k$ (Appendix~\ref{app:spec-proofs})\\
$q$ & $1-m/N$\\
$r(n)$ & bandwidth ratio $\beta_{\mathrm{dedup}}(n)/\beta_{\mathrm{slot}}(n)$\\
$S(n)$ & saving of deduplication, $\tau_{\mathrm{slot}}(n)-\tau_{\mathrm{dedup}}(n)$~\eqref{eq:saving-ratio}\\
$S_{l,t}$ & routing set: the $m$ experts layer $l$ selects for the token at position $t$\\
$t$ & token position\\
$t_0$ & time outside the pass's memory reads\\
$t_{\mathrm{op}}$ & fixed time of a fit on chosen ids\\
$\bar t_{p,l}(n)$ & operation time of layer $l$ with kernel $p$, averaged over windows of recorded routing: the fit at each window's bytes in the prediction, the measured median in the measured-time version~\eqref{eq:cond}\\
$T_a(k)$ & throughput, $G_a(k)/V(k)$ tokens per second\\
$T_{\mathrm{tr}}$ & trace length\\
$U$ & distinct experts of a pass in the benchmark\\
$U_l(n)$ & mean number of distinct experts that $n$ consecutive tokens select in layer $l$\\
$\bar u(n)$ & $\Eu(n)/\Es(n)$, the share of the per-slot bytes that deduplication requests\\
$V(k)$ & step time of a step drafting $k$ tokens\\
$\Vs$, $\Vu$ & bytes of a step with each kernel, the one-bandwidth case (Appendix~\ref{app:spec-proofs})\\
$w_l$, $W$ & bytes of one expert of layer $l$; $W=\sum_l w_l$\\
$X_{l,t}(n)$ & number of distinct experts that $n$ consecutive tokens select in layer $l$, from position $t$\\
\midrule
$\alpha_{k,i}$ & infimum of the acceptances at which draft length $k$ is faster than $i$, 1 if none (Appendix~\ref{app:spec-proofs})\\
$\beta_D$ & bandwidth of the bytes read once per pass\\
$\beta_p(n)$ & effective bandwidth of kernel $p$: its requested expert bytes over its time\\
$\delta$ & bound on the relative error of the predicted step time\\
$\Delta D_k$, $\Delta\phi_k$ & $D(k+1)-D(1)$; $\phi_{\mathrm{dedup}}(k+1)-\phi(1)$\\
$\varepsilon$ & bound on the relative bandwidth error, measured per engine (Section~\ref{sec:dedup-beta})\\
$\rho_l(n)$, $\hat\rho_l(n)$ & reuse of $n$ consecutive tokens~\eqref{eq:reuse}; its estimate on traces\\
$\rbar_k$ & byte-weighted mean reuse of the $k$ drafted tokens~\eqref{eq:rbar}\\
$\rho^*_k(a)$ & reuse threshold: the least $\rbar_k$ at which draft length $k$ is faster than plain decoding~\eqref{eq:rho-star}\\
$\tau_p(n)$ & verify time of a pass of $n$ tokens with kernel $p$~\eqref{eq:tau}\\
$\phi_p(n)$ & fixed work per pass with kernel $p$\\
$\chi$ & cost of a drafted token relative to a target step, in the affine cost of \citet{arXiv:2211.17192}\\
\midrule
$\hat x$ & a quantity computed from the estimated bandwidths, such as $\hat M^{\mathrm{e}}_p$, $\hat V$, $\hat T_a$ and $\hat k^*$\\
\bottomrule
\end{longtable}
\addtocounter{table}{-1}}%
\section{Proofs}
\label{app:spec-proofs}
\begingroup\sloppy

This appendix derives each result of Section~\ref{sec:spec-theory} under its exact conditions. Each paragraph starts from its assumptions, and its heading names the equation of Section~\ref{sec:spec-theory} it derives. A result with no equation there gets a numbered display here. Acceptance is geometric at rate $a\in[0,1)$, and a step with draft length $k$ costs $V(k)>0$. A result that names no kernel holds for every positive cost, measured times included. The results for a deduplicating engine (Appendix~\ref{app:spec-dedup-sec}) assume that the $k+1$ verified tokens are routed like $k+1$ consecutive tokens of a stationary routing process. They are derived in the step time~\eqref{eq:tau}, with the bandwidth of each kernel at each pass size. The byte cost $\Vu(k)=D+\Eu(k+1)+\kv(k)$, with one bandwidth for every byte, is their special case of Appendix~\ref{app:spec-cond-bytes}. Notation is that of Section~\ref{sec:setting} and Section~\ref{sec:spec-theory}. Throughout, $P_k(x)=1+x+\dots+x^k$, so $G_a(k)=P_k(a)$ for $a\in[0,1)$, and every cost is positive. For each result proved in Lean, we name the module of the accompanying Lean package \texttt{MoESpec} and its theorems at the step they prove. We also say which steps, if any, Lean does not cover. The package builds with Lean~4.34.1 against a pinned Mathlib. Every cited theorem uses only the axioms \texttt{propext}, \texttt{Classical.choice} and \texttt{Quot.sound}. The probability models fix the start of the window and live on a finite probability space. The other Lean statements have been compared line by line with the paper's statements. Two groups have not yet been compared in this way: the reuse threshold in time (module \texttt{MoESpec.TimeThreshold}) and the error bounds~\eqref{eq:e1} to~\eqref{eq:e5} (module \texttt{MoESpec.ErrBounds}). Where the only Lean proof of a result for a deduplicating engine is stated for the byte cost, we say so. In that case the step from the byte cost to the step time is by hand. Exact-arithmetic brute-force checks on small cases, which we do not list, also cover four groups of results. These are the speedup condition, the one-bandwidth (byte) forms of the results for a deduplicating engine, the staircase, and the step points under an increment condition. The time forms of the results for a deduplicating engine and the error bounds have no such checks. The Lean models fix the start of the window of positions, and stationarity enters only to drop it. They work on a finite probability space of traces. This space covers every routing process on traces of length $T_{\mathrm{tr}}$, since the set of such traces is finite.

\subsection{Speedup condition}

\paragraph{Speedup condition~\eqref{eq:speedup-cond}.}
\label{app:spec-cond}
Here $\kmax\ge1$, $V:\{0,\dots,\kmax\}\to(0,\infty)$ is any cost, $a\in[0,1)$ and $k\in\{1,\dots,\kmax\}$. No property of $V$ beyond positivity is used. The geometric sum gives
\begin{equation}
G_a(0)=1,\qquad G_a(k)-1=a+a^2+\dots+a^k .
\label{eq:app-G}
\end{equation}
Since $V(0),V(k)>0$, multiplying by $V(0)V(k)>0$ turns the comparison of throughputs into one of costs,
\begin{equation}
T_a(k)>T_a(0)\iff\frac{G_a(k)}{V(k)}>\frac{1}{V(0)}\iff G_a(k)\,V(0)>V(k)\iff V(k)-V(0)<\bigl(G_a(k)-1\bigr)V(0),
\label{eq:app-cond-k}
\end{equation}
which by~\eqref{eq:app-G} is~\eqref{eq:speedup-cond} (Lean: \texttt{MoESpec.SpeedupCond}, \texttt{faster\_at\_k}). Speculation gives a speedup at $a$ when $k^*(a)\ge1$, and
\begin{equation}
k^*(a)\ge1\iff T_a(k)>T_a(0)\ \text{for some}\ k\in\{1,\dots,\kmax\}.
\label{eq:app-speedup-some-k}
\end{equation}
If $k^*(a)\ge1$, then $k^*(a)$ is a maximizer and $0$ is not, because the smallest maximizer is positive. So $T_a(k^*(a))>T_a(0)$. Conversely, if some $k\ge1$ has $T_a(k)>T_a(0)$, then $0$ is not a maximizer and $k^*(a)\ge1$. With~\eqref{eq:app-cond-k}, speculation therefore gives a speedup at $a$ if and only if $V(k)-V(0)<(G_a(k)-1)V(0)$ for some $k\in\{1,\dots,\kmax\}$. In Lean this is \texttt{MoESpec.SpeedupCondIff}, \texttt{faster\_iff}, whose ``faster'' is the right side of~\eqref{eq:app-speedup-some-k}. The equivalence~\eqref{eq:app-speedup-some-k} is by hand. For $k=1$, $G_a(1)-1=a$, so~\eqref{eq:app-cond-k} reads $V(1)-V(0)<aV(0)$. We divide by $V(0)>0$ and repeat the computation with equality for the second case. This gives
\begin{equation}
T_a(1)>T_a(0)\iff a>a_0,\qquad T_a(1)=T_a(0)\iff a=a_0,\qquad a_0=\frac{V(1)}{V(0)}-1
\label{eq:app-a0}
\end{equation}
(Lean: \texttt{MoESpec.FirstTokenThreshold}, \texttt{thresh} and \texttt{thresh\_eq}).

\subsection{Deduplicating engines}
\label{app:spec-dedup-sec}

\paragraph{Union bytes~\eqref{eq:union-reuse}, \eqref{eq:Eu-rbar}.}
\label{app:spec-cond-union}
Here the routing process is stationary, with sets of $m\ge1$ experts and $w_l>0$. Nothing about a kernel or a bandwidth enters. Fix $l$, $t$ and $n\ge1$, and let $R=S_{l,t}\cup\dots\cup S_{l,t+n-2}$, which is empty for $n=1$. Since $|S_{l,t+n-1}|=m$,
\begin{equation}
X_{l,t}(n)=|R\cup S_{l,t+n-1}|=|R|+|S_{l,t+n-1}|-|R\cap S_{l,t+n-1}|=X_{l,t}(n-1)+m-|S_{l,t+n-1}\cap R|.
\label{eq:app-union-step}
\end{equation}
We take expectations with both windows starting at $t$. Stationarity makes the result independent of $t$. Using $\E|S_{l,t+n-1}\cap R|=m\rho_l(n)$ by~\eqref{eq:reuse}, we get
\begin{equation}
U_l(n)-U_l(n-1)=m\bigl(1-\rho_l(n)\bigr)
\label{eq:app-union-inc}
\end{equation}
(Lean: \texttt{MoESpec.ReuseUnion}, \texttt{increment}). Summing from $U_l(0)=0$ gives~\eqref{eq:union-reuse} (\texttt{telescope}). Multiplying by $w_l$ and summing over $l$ gives~\eqref{eq:class-gap} and
\begin{equation}
\Eu(n)=m\sum_lw_l\sum_{j\le n}\bigl(1-\rho_l(j)\bigr)
\label{eq:app-Eu}
\end{equation}
(\texttt{MoESpec.ClassGap}, \texttt{union\_eq} and \texttt{gap}). At $n=k+1$ the term $j=1$ is $m\sum_lw_l=mW$, since $\rho_l(1)=0$. By the definition~\eqref{eq:rbar} of $\rbar_k$, the remaining terms give
\begin{equation}
\Eu(k+1)=mW+m\sum_lw_l\sum_{j=2}^{k+1}\bigl(1-\rho_l(j)\bigr)=mW+mWk-mWk\,\rbar_k ,
\label{eq:app-Eu-split}
\end{equation}
which is~\eqref{eq:Eu-rbar}. This step is by hand. In Lean it appears only inside the byte cost (\texttt{MoESpec.SpeedupCond}, lemma \texttt{Vu\_diff}). The union of $k+1$ sets contains the first set's $m$ experts, so $X_{l,t}(k+1)\ge m$. With $w_l>0$, this gives
\begin{equation}
\Eu(k+1)\ge mW>0
\label{eq:app-Eu-pos}
\end{equation}
(Lean: \texttt{MoESpec.Positivity}, \texttt{positivity}, whose second and third parts are~\eqref{eq:app-Eu-pos}).

\paragraph{Step time~\eqref{eq:dV-time}.}
\label{app:spec-dedup-time}
Here the engine deduplicates with $f_l=1$, the draft model's and the host's time $h$ is zero, and the routing process is as in the previous paragraph. By~\eqref{eq:tau}, $t_0\ge0$, $\beta_D>0$, $D\ge0$ and $\kv\ge0$. The fixed work $\phi(1)$ and $\phi_{\mathrm{dedup}}(k+1)$ is non-negative. The bandwidths $\beta(1)$ and $B_k=\beta_{\mathrm{dedup}}(k+1)$ are positive, and each is a function of the pass size alone. A one-token pass reads $\Eu(1)=mW$ expert bytes. So with~\eqref{eq:Eu-rbar}, the step times are, for $k\in\{1,\dots,\kmax\}$,
\begin{equation}
\begin{aligned}
V(0)&=\tau(1)=t_0+\frac{D(1)}{\beta_D}+\frac{mW}{\beta(1)}+\phi(1),\\
V(k)&=\tau_{\mathrm{dedup}}(k+1)=t_0+\frac{D(k+1)}{\beta_D}+\frac{mW+mWk(1-\rbar_k)}{B_k}+\phi_{\mathrm{dedup}}(k+1).
\end{aligned}
\label{eq:app-time-costs}
\end{equation}
When we subtract, $t_0$ cancels. With $\Delta D_k$ and $\Delta\phi_k$ as in Section~\ref{sec:spec-dedup}, the difference is~\eqref{eq:dV-time}. Every term of~\eqref{eq:app-time-costs} is non-negative. By~\eqref{eq:app-Eu-pos}, the expert term of $V(k)$ is at least $mW/B_k$, so
\begin{equation}
V(0)\ge\frac{mW}{\beta(1)}>0,\qquad V(k)\ge\frac{\Eu(k+1)}{B_k}\ge\frac{mW}{B_k}>0 .
\label{eq:app-time-pos}
\end{equation}
The byte inequality~\eqref{eq:app-Eu-pos} is in Lean. The step from it to~\eqref{eq:app-time-pos} is by hand, and the Lean theorems in time take positivity of $V$ as a hypothesis. For the byte cost, \texttt{MoESpec.Positivity}, \texttt{Vu\_pos} and \texttt{U1\_holds} prove $\Vu(k)\ge\Eu(k+1)\ge mW>0$ from the setting and discharge that hypothesis. A non-negative $h$ would enter~\eqref{eq:app-time-costs} as extra fixed work, $h(0)$ in $\phi(1)$ and $h(k)$ in $\phi_{\mathrm{dedup}}(k+1)$. It would leave every step below unchanged.

\paragraph{Reuse threshold~\eqref{eq:rho-star}.}
\label{app:spec-cond-time}
Here the step times are~\eqref{eq:app-time-costs} under the assumptions of the previous paragraph. Also $a\in[0,1)$, $k\in\{1,\dots,\kmax\}$ and $\rho^*_k(a)$ is~\eqref{eq:rho-star}. Write
\begin{equation}
Y=\bigl(G_a(k)-1\bigr)\tau(1)-\frac{\Delta D_k}{\beta_D}-\Delta\phi_k
\label{eq:app-Y}
\end{equation}
for the bracket of~\eqref{eq:rho-star}. By~\eqref{eq:dV-time},
\begin{equation}
V(k)-V(0)-\bigl(G_a(k)-1\bigr)V(0)=\frac{mW+mWk(1-\rbar_k)}{B_k}-\frac{mW}{\beta(1)}-Y=\frac{mWk}{B_k}\bigl(\rho^*_k(a)-\rbar_k\bigr),
\label{eq:app-gap}
\end{equation}
where the last equality expands $\frac{mWk}{B_k}\rho^*_k(a)=\frac{mWk}{B_k}-Y-\frac{mW}{\beta(1)}+\frac{mW}{B_k}$ (Lean: \texttt{MoESpec.TimeThreshold}, \texttt{gap\_identity}). That theorem needs only $m,W,\beta(1),B_k>0$ and $k\ge1$. Since $V>0$ by~\eqref{eq:app-time-pos}, \eqref{eq:app-cond-k} applies. So $T_a(k)>T_a(0)$ if and only if the left side of~\eqref{eq:app-gap} is negative. Since $mWk/B_k>0$, this gives
\begin{equation}
T_a(k)>T_a(0)\iff\rbar_k>\rho^*_k(a)
\label{eq:app-time-iff}
\end{equation}
(\texttt{faster\_time}, with positivity of $V$ as a hypothesis). By~\eqref{eq:app-speedup-some-k}, speculation gives a speedup at $a$ if and only if $\rbar_k>\rho^*_k(a)$ for some $k\in\{1,\dots,\kmax\}$. Each $k$ has its own $B_k$, $\Delta D_k$ and $\Delta\phi_k$. This union over $k$ is by hand in time form, and in Lean only for the byte cost (\texttt{MoESpec.SpeedupCondIff}, \texttt{faster\_iff\_reuse}). Now fix $t_0$, $\beta_D$, $D(k+1)$, $\phi_{\mathrm{dedup}}(k+1)$ and $B_k$. Then only the expert term of $V(k)$ in~\eqref{eq:app-time-costs} depends on $\rbar_k$. Its numerator $mW+mWk(1-\rbar_k)$ falls as $\rbar_k$ rises. Since $B_k>0$, $V(k)$ is non-increasing in $\rbar_k$, and $V(0)$ does not depend on it:
\begin{equation}
\rbar_k\le\rbar'_k\;\Longrightarrow\;V(k)\big|_{\rbar'_k}\le V(k)\big|_{\rbar_k}
\label{eq:app-time-mono}
\end{equation}
(\texttt{tauk\_mono}). Suppose moreover that $\beta_D=B_k=\beta(1)=\beta>0$ and $t_0=\phi(1)=\phi_{\mathrm{dedup}}(k+1)=0$. Then $\tau(1)=(D(1)+mW)/\beta$, $B_k/\beta(1)=1$ and $B_kY/(mWk)=[(G_a(k)-1)(D(1)+mW)-(D(k+1)-D(1))]/(mWk)$. Substituting into~\eqref{eq:rho-star} gives
\begin{equation}
\rho^*_k(a)=1-\frac{\bigl(G_a(k)-1\bigr)\bigl(D(1)+mW\bigr)-\bigl(D(k+1)-D(1)\bigr)}{mWk}
\label{eq:app-time-bytes}
\end{equation}
(\texttt{byte\_form}). This is the one-bandwidth case of Appendix~\ref{app:spec-cond-bytes}. Finally, for any positive cost, $T_a(0)=G_a(0)/V(0)=1/V(0)$ and $V(k)/V(0)=1+(V(k)-V(0))/V(0)$, so
\begin{equation}
\frac{T_a(k)}{T_a(0)}=\frac{G_a(k)\,V(0)}{V(k)}=\frac{G_a(k)}{1+\bigl(V(k)-V(0)\bigr)/V(0)}
\label{eq:app-time-ratio}
\end{equation}
(\texttt{speedup\_ratio}). Substituting~\eqref{eq:dV-time} for $V(k)-V(0)$ gives~\eqref{eq:speedup-dedup}. The substitution is by hand.

\paragraph{One draft.}
\label{app:spec-time-k1}
Here the assumptions are those of the previous paragraph with $k=1$. Then $G_a(1)-1=a$, $\rbar_1=\sum_lw_l\rho_l(2)/W$ by~\eqref{eq:rbar}, and $\Eu(2)=mW(2-\rbar_1)$ by~\eqref{eq:Eu-rbar}. We substitute these into~\eqref{eq:rho-star}. We also substitute them, with~\eqref{eq:dV-time} and $V(0)=\tau(1)$, into $a_0=(V(1)-V(0))/V(0)$ of~\eqref{eq:app-a0}. This gives
\begin{equation}
\rho^*_1(a)=1-\frac{B_1}{mW}\Bigl[a\,\tau(1)-\frac{\Delta D_1}{\beta_D}-\Delta\phi_1\Bigr]-\Bigl(\frac{B_1}{\beta(1)}-1\Bigr),\qquad a_0=\frac{1}{\tau(1)}\Bigl[\frac{\Delta D_1}{\beta_D}+\Delta\phi_1+\frac{mW(2-\rbar_1)}{B_1}-\frac{mW}{\beta(1)}\Bigr].
\label{eq:app-time-k1}
\end{equation}
By~\eqref{eq:app-time-iff} and~\eqref{eq:app-a0}, one draft is faster than none exactly when $\rbar_1>\rho^*_1(a)$, equivalently when $a>a_0$. Both equivalences are in Lean in time form. The first is \texttt{MoESpec.TimeThreshold}, \texttt{faster\_time} at $k=1$. The second is \texttt{MoESpec.FirstTokenThreshold}, \texttt{thresh}, for any positive cost. The explicit forms~\eqref{eq:app-time-k1} are substitutions by hand. Lean states the explicit $a_0$ only for the byte cost (\texttt{MoESpec.SpeedupCond}, \texttt{a0\_union}).

\paragraph{Threshold movement.}
\label{app:spec-thresh-move}
Here the assumptions are those of the reuse threshold in time, with $k\ge1$ fixed. Moving $a$ changes only $G_a(k)$ in~\eqref{eq:rho-star}, because no other quantity in it depends on $a$. The bandwidths and the fixed work are functions of the pass size alone. Moving $D$ to $D'$ changes only $\tau(1)$, by $(D'-D)/\beta_D$, since $\Delta D_k=\kv(k)-\kv(0)$ does not involve $D$. This holds provided the bandwidths and the fixed work do not change with $D$. So, for $a,a'\in[0,1)$,
\begin{equation}
\rho^*_k(a)-\rho^*_k(a')=\frac{B_k\,\tau(1)}{mWk}\bigl(G_{a'}(k)-G_a(k)\bigr),\qquad\rho^*_k(a)\big|_{D}-\rho^*_k(a)\big|_{D'}=\frac{B_k\,\bigl(G_a(k)-1\bigr)}{mWk\,\beta_D}\,(D'-D).
\label{eq:app-time-move}
\end{equation}
For $0\le a<a'<1$, every term of $G_a(k)-1=a+\dots+a^k$ is at most the matching term at $a'$, and the first is smaller. So $G_{a'}(k)>G_a(k)$. With $B_k,\tau(1)>0$ by~\eqref{eq:app-time-pos}, $\rho^*_k(a)$ is strictly decreasing in $a$ on $[0,1)$. For $a>0$, $G_a(k)-1>0$ and $\beta_D>0$, so $\rho^*_k(a)$ is strictly decreasing in $D$. At $a=0$ it does not depend on $D$. These steps are by hand. Lean proves the same three statements only for the byte form~\eqref{eq:rho-star-bytes} (\texttt{MoESpec.ThresholdMonotone}, \texttt{rhoStar\_lt} and \texttt{U2\_holds}).

\paragraph{More reuse.}
\label{app:spec-more-reuse}
Here two stationary routing processes, a prime marking the second, have reuses $\rho_l(j)\le\rho'_l(j)$ for every $l$ and $j=2,\dots,\kmax+1$. They have the same $L$, $N$, $m$, $w_l$, $D$, $\kv$ and $a$. Their step times are~\eqref{eq:app-time-costs} under the assumptions of the step-time paragraph. They have the same $t_0$, $\beta_D$, $\phi(1)$ and $\beta(1)$ and, for every $k$, the same $B_k$ and $\phi_{\mathrm{dedup}}(k+1)$. That is, the deduplicating bandwidth and fixed work at a fixed pass size do not depend on which experts repeat. Since $w_l>0$, the definition~\eqref{eq:rbar} gives the second relation below. The first holds because $\tau(1)$ involves no reuse:
\begin{equation}
V'(0)=V(0),\qquad\rbar_k\le\rbar'_k\quad(k=1,\dots,\kmax),
\label{eq:app-more-reuse}
\end{equation}
so by~\eqref{eq:app-time-mono}, $V'(k)\le V(k)$ for every $k$ (\texttt{MoESpec.TimeThreshold}, \texttt{tauk\_mono}). The comparison of $\rbar_k$ is by hand. Every input of $\rho^*_k(a)$ in~\eqref{eq:rho-star} is shared by the two processes. Suppose that speculation gives a speedup at $a$ under the first process. Then by~\eqref{eq:app-time-iff}, some $k\in\{1,\dots,\kmax\}$ has $\rbar_k>\rho^*_k(a)$, and by~\eqref{eq:app-more-reuse}
\begin{equation}
\rbar'_k\ge\rbar_k>\rho^*_k(a),
\label{eq:app-more-reuse-faster}
\end{equation}
and by~\eqref{eq:app-time-iff} for the second process ($V'>0$ by~\eqref{eq:app-time-pos}), speculation gives a speedup at $a$ under the second. Each use of~\eqref{eq:app-time-iff} is \texttt{faster\_time}. The chaining is by hand. With $V'(0)=V(0)$ and $V'(k)\le V(k)$, \eqref{eq:app-time-ratio} also shows that the speedup $T_a(k)/T_a(0)$ does not fall. This step is by hand. The one-bandwidth case is in Lean for the byte cost (\texttt{MoESpec.MoreReuse}, \texttt{cost\_zero}, \texttt{cost\_le} and \texttt{faster\_mono}). If $B_k$ or $\phi_{\mathrm{dedup}}(k+1)$ changes with the routing at fixed pass size, then $\rho^*_k(a)$ may change too. In that case the argument above does not apply.

\paragraph{One bandwidth~\eqref{eq:rho-star-bytes}.}
\label{app:spec-cond-bytes}
Here $\beta_D=B_k=\beta(1)=\beta>0$ for every $k$ and $t_0=\phi(1)=\phi_{\mathrm{dedup}}(k+1)=0$. This is an admissible instance of~\eqref{eq:app-time-costs}. With $D(n)=D+\kv(n-1)$, \eqref{eq:app-time-costs} becomes $V(k)=\Vu(k)/\beta$ with $\Vu(k)=D+\Eu(k+1)+\kv(k)$. So every throughput is multiplied by $\beta$, and no comparison of throughputs changes. With $D(1)=D+\kv(0)$ and $D(k+1)-D(1)=\kv(k)-\kv(0)$, \eqref{eq:app-time-bytes} is~\eqref{eq:rho-star-bytes}, with $\Vu(0)=D+mW+\kv(0)$ (\texttt{MoESpec.TimeThreshold}, \texttt{byte\_form}). At $k=1$, \eqref{eq:app-time-k1} becomes
\begin{equation}
\rho^*_1(a)=1-\frac{a\Vu(0)-\bigl(\kv(1)-\kv(0)\bigr)}{mW},\qquad a_0=\frac{m\sum_lw_l\bigl(1-\rho_l(2)\bigr)+\kv(1)-\kv(0)}{\Vu(0)} ,
\label{eq:app-bytes-k1}
\end{equation}
using $mW(1-\rbar_1)=m\sum_lw_l(1-\rho_l(2))$. In this case, Lean also proves the results of the previous paragraphs directly from the routing process, in the byte cost. These results are the threshold and the speedup at some $k$ (\texttt{MoESpec.SpeedupCond}, \texttt{faster\_reuse}; \texttt{MoESpec.SpeedupCondIff}, \texttt{faster\_iff\_reuse}; \texttt{MoESpec.Positivity}, \texttt{U1\_holds}, which also discharges positivity). They also include the explicit $a_0$ (\texttt{MoESpec.SpeedupCond}, \texttt{a0\_union}), the movement in $a$ and $D$ (\texttt{MoESpec.ThresholdMonotone}) and more reuse (\texttt{MoESpec.MoreReuse}).

\paragraph{No single threshold.}
\label{app:no-single}
Here two stationary routing processes share every quantity that the paragraph on more reuse holds fixed, with no order between their reuses. Then $\rho^*_k(a)$ is the same for both. So by~\eqref{eq:app-time-iff}, two routing processes with equal $(\rbar_1,\dots,\rbar_{\kmax})$ both give a speedup or neither does. This step is by hand from \texttt{faster\_time}. Lean proves it for the byte cost (\texttt{MoESpec.NoSingleThreshold}, \texttt{U3a\_holds}). With $\kmax\ge2$, no single $\rbar_k$ suffices. The example below uses one bandwidth with $t_0=\phi=0$. This is an admissible instance of the time model (Appendix~\ref{app:spec-cond-bytes}), so the example shows the same for the step time. Take $N=3$, $L=1$, $m=1$, $w_1=1$, $D=0$, $\kv\equiv0$, $\kmax=2$ and $a=13/20$. Then $\rho^*_1=7/20$ and $\rho^*_2=371/800$ by~\eqref{eq:rho-star-bytes}. Each process below repeats one pattern of experts, started at a uniformly random phase, so it is stationary.
\begin{center}
\begin{tabular}{lcccc}
\toprule
Pattern & Period & $\rbar_1$ & $\rbar_2$ & Speedup\\
\midrule
\texttt{01} & 2 & 0 & 1/2 & yes ($k=2$)\\
\texttt{00012} & 5 & 2/5 & 2/5 & yes ($k=1$)\\
\texttt{00102} & 5 & 1/5 & 2/5 & no\\
\bottomrule
\end{tabular}
\end{center}
For $k=1$, pattern \texttt{01} gives a speedup with a lower $\rbar_1$ than \texttt{00102}, which does not. For $k=2$, \texttt{00012} gives a speedup with the same $\rbar_2$ as \texttt{00102}. So no rule ``$\rbar_k>c$'' or ``$\rbar_k\ge c$'' decides for either $k$. Lean: \texttt{MoESpec.NoSingleThreshold}, \texttt{U3b\_holds} and \texttt{U3\_holds}, for the byte cost. The example has $D=0$; we built none with $D>0$.

\subsection{Staircase}
\label{app:spec-stair-sec}

\paragraph{Monotonicity and step points.}
\label{app:spec-stair}
Here $\kmax\ge1$, $V:\{0,\dots,\kmax\}\to(0,\infty)$ is any cost and $a\in[0,1)$. For $0\le i<k\le\kmax$, let $\mathcal J_{k,i}=\{a\in[0,1):T_a(k)>T_a(i)\}$, and let $\alpha_{k,i}$ be its infimum, taken as 1 when the set is empty.

The ratio $G_a(k)/G_a(i)$ rises with $a$. For $i<k$,
\begin{equation}
P_k(x)=P_i(x)+x^{i+1}P_{k-i-1}(x).
\label{eq:app-P-split}
\end{equation}
For $0\le x\le y$, termwise, since $x^{i+1}y^s=(xy)^sx^{i+1-s}\le(xy)^sy^{i+1-s}$ for $s\le i$,
\begin{equation}
x^{i+1}P_i(y)\le y^{i+1}P_i(x),
\label{eq:app-key}
\end{equation}
and $P_{k-i-1}(x)\le P_{k-i-1}(y)$. Hence, by~\eqref{eq:app-P-split} at $x$ and at $y$,
\begin{equation}
P_k(x)P_i(y)=P_i(x)P_i(y)+x^{i+1}P_{k-i-1}(x)P_i(y)\le P_i(x)P_i(y)+y^{i+1}P_{k-i-1}(y)P_i(x)=P_k(y)P_i(x),
\label{eq:app-P-chain}
\end{equation}
that is, for $0\le i<k\le\kmax$ and $0\le x\le y<1$,
\begin{equation}
\frac{G_x(k)}{G_x(i)}\le\frac{G_y(k)}{G_y(i)}
\label{eq:app-stair-ratio}
\end{equation}
(Lean: \texttt{MoESpec.Staircase}, \texttt{ratio\_mono}). Since
\begin{equation}
T_a(k)>T_a(i)\iff\frac{G_a(k)}{G_a(i)}>\frac{V(k)}{V(i)},
\label{eq:app-stair-T}
\end{equation}
the set $\mathcal J_{k,i}$ contains, with any $a$, every $a'\in[a,1)$. That is, it is closed upward in $[0,1)$ (\texttt{upset}). So it contains $(\alpha_{k,i},1)$ and misses $[0,\alpha_{k,i})$.

The best draft length rises with $a$. Let $a\le a'$ in $[0,1)$ and $k^*(a)\ge j$, so no $i<j$ maximizes $T_a$. Take $i<j$ and a maximizer $k_0$ of $T_a$. Then $k_0\ge j>i$ and $T_a(i)<T_a(k_0)$, so $a\in\mathcal J_{k_0,i}$. Hence $a'\in\mathcal J_{k_0,i}$, that is, $T_{a'}(k_0)>T_{a'}(i)$, and $i$ does not maximize $T_{a'}$. So $k^*(a')\ge j$. As this holds for every $j$,
\begin{equation}
0\le a\le a'<1\;\Longrightarrow\;k^*(a)\le k^*(a')
\label{eq:app-stair-mono}
\end{equation}
(\texttt{kstar\_mono}).

The step points are exact. For $j\in\{1,\dots,\kmax\}$ and $a\in[0,1)$,
\begin{equation}
k^*(a)\ge j\iff\text{every } i\in\{0,\dots,j-1\} \text{ has } T_a(i)<T_a(k) \text{ for some } k\in\{j,\dots,\kmax\}.
\label{eq:app-stair-ge}
\end{equation}
``If'': each $i<j$ has lower throughput than some $k\ge j$. So no $i<j$ is a maximizer, and the smallest maximizer is at least $j$. ``Only if'': the maximizer $k^*(a)\ge j$ has higher throughput than each $i<j$, since none of them is a maximizer. In Lean this is \texttt{ge\_iff}, which states $k^*(a)\ge j$ as ``no $i<j$ maximizes $T_a$''. The equivalence of that statement with $k^*(a)\ge j$ for the smallest maximizer is by hand. By~\eqref{eq:app-stair-ge},
\begin{equation}
\{a\in[0,1):k^*(a)\ge j\}=\bigcap_{0\le i<j}\ \bigcup_{j\le k\le\kmax}\mathcal J_{k,i}.
\label{eq:app-stair-sets}
\end{equation}
A union of upward-closed sets is upward closed, with infimum the least infimum. A finite intersection of them is upward closed, with infimum the largest. So the set~\eqref{eq:app-stair-sets} contains every $a$ above $\max_{i<j}\min_{k\ge j}\alpha_{k,i}$ and none below it. This number is its infimum, the step point:
\begin{equation}
a_j=\max_{0\le i<j}\ \min_{j\le k\le\kmax}\alpha_{k,i},
\label{eq:app-aj}
\end{equation}
that is, $k^*(a)\ge j$ for every $a\in(a_j,1)$ and $k^*(a)<j$ for every $a\in[0,a_j)$ (Lean: \texttt{MoESpec.StepPoints}, \texttt{step\_maxmin}, \texttt{ge\_of\_gt\_maxmin} and \texttt{not\_ge\_of\_lt\_maxmin}). For $j=1$ only $i=0$ occurs, so $a_1=\min_k\alpha_{k,0}$ (\texttt{step\_one}). So speculation gives a speedup at every $a\in(a_1,1)$ and at no $a\in[0,a_1)$.

\paragraph{Jumps.}
\label{app:stair-jump}
The statement of Section~\ref{sec:spec-stair} that the staircase can jump by more than one length rests on a two-expert example. Take $L=1$, $N=2$, $m=1$, $w_1=1$, $D=0$, $\kv\equiv0$ and $\kmax=2$. Tokens that alternate between the two experts with a uniformly random phase then give $\Vu=(1,2,2)$. This is the step time~\eqref{eq:tau} with one bandwidth $\beta=1$ and $t_0=\phi=0$, an admissible instance of the time model. Then $k^*(1/2)=0$ and $k^*(9/10)=2$, so this staircase jumps from 0 to 2 (\texttt{MoESpec.UnionJump}, \texttt{union\_half}; \texttt{MoESpec.UnionJump2}, \texttt{union\_half\_U} and \texttt{alt\_stair\_U}). That $k^*(a)\ne1$ for every $a$ is checked by hand.

\paragraph{Convex step time~\eqref{eq:stair-step}.}
\label{app:spec-stair-inc}
Here $\kmax\ge1$, $V:\{0,\dots,\kmax\}\to(0,\infty)$ is any cost, $a\in[0,1)$ and $j\in\{1,\dots,\kmax\}$, and
\begin{equation}
h_j(a)=\frac{a^j}{G_a(j-1)}.
\label{eq:app-hj}
\end{equation}
The first part assumes $c_{i+1}\ge a\,c_i$ for every $i\in\{1,\dots,\kmax-1\}$. The second part assumes $0\le c_1\le\dots\le c_{\kmax}$, that is, $V$ convex and non-decreasing.

Once $T_a$ stops rising it never rises again. Write $P_i=P_i(a)$. Since $V(i+1)=V(i)+c_{i+1}$ and $P_{i+1}=P_i+a^{i+1}$,
\begin{equation}
T_a(i+1)\le T_a(i)\iff\bigl(P_i+a^{i+1}\bigr)V(i)\le P_iV(i+1)\iff a^{i+1}V(i)\le P_i\,c_{i+1}.
\label{eq:app-fall}
\end{equation}
If $T_a(i+1)\le T_a(i)$, then by~\eqref{eq:app-fall}, $a\ge0$, $P_{i+1}>0$ and the hypothesis $c_{i+2}\ge a\,c_{i+1}$,
\begin{equation}
a^{i+2}V(i+1)=a\bigl(a^{i+1}V(i)+a^{i+1}c_{i+1}\bigr)\le a\bigl(P_i+a^{i+1}\bigr)c_{i+1}=P_{i+1}\,a\,c_{i+1}\le P_{i+1}\,c_{i+2},
\label{eq:app-propagate}
\end{equation}
which is $T_a(i+2)\le T_a(i+1)$ by~\eqref{eq:app-fall} at $i+1$ (Lean: \texttt{MoESpec.Staircase}, \texttt{propagate}). ``If'': suppose $T_a(j)>T_a(j-1)$. Then $T_a$ rose at every step before $j$, because a fall would propagate to the step from $j-1$ to $j$. So every $i<j$ has $T_a(i)<T_a(j)$, and $k^*(a)\ge j$. ``Only if'': suppose $T_a(j)\le T_a(j-1)$. Then $T_a(k)\le T_a(j-1)$ for every $k\ge j$, and a maximizer over $0,\dots,j-1$ is a global maximizer below $j$ (\texttt{uni\_iff}). Negating~\eqref{eq:app-fall} at $i=j-1$ shows that $T_a(j)>T_a(j-1)$ if and only if $(P_{j-1}+a^j)V(j-1)>P_{j-1}V(j)$, that is, $a^jV(j-1)>G_a(j-1)\,c_j$ (\texttt{step\_form}). So under $c_{i+1}\ge a\,c_i$ for every $i\in\{1,\dots,\kmax-1\}$,
\begin{equation}
k^*(a)\ge j\iff T_a(j)>T_a(j-1)\iff a^j\,V(j-1)>G_a(j-1)\,c_j .
\label{eq:app-inc-i}
\end{equation}

Under $0\le c_1\le\dots\le c_{\kmax}$, $0\le c_i\le c_{i+1}$ gives $c_{i+1}\ge c_i\ge a\,c_i$. So the hypothesis of~\eqref{eq:app-inc-i} holds at every $a$, and~\eqref{eq:app-inc-i} is~\eqref{eq:stair-step}. Dividing its last inequality by $G_a(j-1)V(j-1)>0$ gives the first statement below. For $0\le x<y<1$, $x^jP_{j-1}(y)<y^jP_{j-1}(x)$ termwise, since each term satisfies $x^jy^s\le y^jx^s$ for $s\le j-1$, strictly at $s=0$. So $h_j$ is strictly increasing on $[0,1)$ (\texttt{h\_strict}), and $\{a:h_j(a)>c_j/V(j-1)\}$ is upward closed with infimum $a_j$. For $j=1$, $h_1(a)=a$, and the infimum of $\{a\in[0,1):a>c_1/V(0)\}$ is $\min(c_1/V(0),1)$ (Lean: \texttt{MoESpec.StepPoints}, \texttt{a1\_uni}). So
\begin{equation}
k^*(a)\ge j\iff h_j(a)>\frac{c_j}{V(j-1)},\qquad a_j=\inf\Bigl\{a\in[0,1):h_j(a)>\frac{c_j}{V(j-1)}\Bigr\},\qquad a_1=\min\Bigl(\frac{c_1}{V(0)},1\Bigr),
\label{eq:app-inc-ii}
\end{equation}
and $a_1$ is $a_0$ when $a_0<1$. The theorem \texttt{MoESpec.StaircaseInfimum}, \texttt{t5}, proves~\eqref{eq:app-inc-i} and~\eqref{eq:app-inc-ii} whole. This includes that the hypothesis of~\eqref{eq:app-inc-ii} implies that of~\eqref{eq:app-inc-i}, and the reading of $a_j$ as the infimum of an upward-closed set.

\endgroup
\section{Error-bound proofs}
\label{app:err-proofs}
\begingroup\sloppy

This appendix derives the bounds of Table~\ref{tab:bounds}. Each paragraph starts from its assumptions, and its heading names the bound it derives. Throughout, the step time is~\eqref{eq:tau}. Each bandwidth in it is estimated as $\hat\beta=\beta(1+e)$ with $|e|\le\varepsilon<1$, one $e$ per bandwidth. Every other term of~\eqref{eq:tau} is exact and non-negative. Each bound is proved in the module \texttt{MoESpec.ErrBounds} of the accompanying Lean package. The proofs use Mathlib at the pinned version and only the standard axioms. We name the theorems at the step they prove. The other Lean results of Appendix~\ref{app:spec-proofs} have been compared line by line with the statements they formalize. These Lean statements have not yet been compared in this way with the statements below. The same holds for the reuse threshold in time (Eqs.~\eqref{eq:app-time-iff} to~\eqref{eq:app-time-ratio}).

\paragraph{Verify time~\eqref{eq:e1}.}
Here $M\ge0$ is the bytes of one memory term of~\eqref{eq:tau} and $\beta>0$ its bandwidth. The estimate moves the term by
\begin{equation}
\frac{M}{\beta(1+e)}-\frac{M}{\beta}=-\frac{M}{\beta}\,\frac{e}{1+e}.
\label{eq:app-err-term}
\end{equation}
Since $1+e\ge1-\varepsilon>0$, cross-multiplying turns $|e|/(1+e)\le\varepsilon/(1-\varepsilon)$ into $|e|(1-\varepsilon)\le\varepsilon(1+e)$. This inequality follows from $|e|\le\varepsilon$ and $-|e|\le e$. So each memory term moves by at most $\varepsilon/(1-\varepsilon)$ of itself (\texttt{time\_err}). By the triangle inequality, the memory part $M_p(n)=D(n)/\beta_D+E_p(n)/\beta_p(n)$ also moves by at most $\varepsilon/(1-\varepsilon)$ of itself (\texttt{time\_err2}). The other terms of $\tau_p(n)$ are exact, so $\hat\tau_p-\tau_p=\hat M_p-M_p$. They are also non-negative, so $M_p(n)\le\tau_p(n)$. Together these give~\eqref{eq:e1}.

\paragraph{Saving~\eqref{eq:e2}.}
Here both kernels read the same bytes $D(n)$ once per pass at the same dense bandwidth $\beta_D$, with the same estimate. By the previous paragraph, $\hat\tau_p-\tau_p=\hat M_p-M_p$ for each kernel. The dense term is the same for both kernels, so
\begin{equation}
\hat S(n)-S(n)=\Bigl(\frac{\Es(n)}{\hat\beta_{\mathrm{slot}}(n)}-\frac{\Es(n)}{\beta_{\mathrm{slot}}(n)}\Bigr)-\Bigl(\frac{E_{\mathrm{dedup}}(n)}{\hat\beta_{\mathrm{dedup}}(n)}-\frac{E_{\mathrm{dedup}}(n)}{\beta_{\mathrm{dedup}}(n)}\Bigr).
\label{eq:app-err-saving}
\end{equation}
Each bracket is one memory term of~\eqref{eq:app-err-term}. So the triangle inequality and the bound of the previous paragraph on each bracket give~\eqref{eq:e2} (\texttt{saving\_err}).

\paragraph{Faster kernel~\eqref{eq:e3}.}
Here $x,y\ge0$ are the true expert times $M^{\mathrm{e}}_p(n)$ of the two kernels and $\phi_x,\phi_y\ge0$ their fixed work. The estimates multiply $x$ and $y$ by factors $u,v=1/(1+e)$. Each factor lies in $[\ell_-,\ell_+]=[1/(1+\varepsilon),1/(1-\varepsilon)]$, with $\ell_-\le1\le\ell_+$. If
\begin{equation}
x\ell_++\phi_x<y\ell_-+\phi_y ,
\label{eq:app-err-margin}
\end{equation}
then $xu+\phi_x\le x\ell_++\phi_x<y\ell_-+\phi_y\le yv+\phi_y$. Also $x+\phi_x\le x\ell_++\phi_x<y\ell_-+\phi_y\le y+\phi_y$. So the predicted and the true times order the kernels alike (\texttt{decision\_agree\_f}). The theorem \texttt{decision\_agree} proves the case $\phi_x=\phi_y=0$ separately, for $x,y>0$. Now take $f_l=1$ and equal fixed work, so that $\phi_x=\phi_y$. This fixed work cancels from~\eqref{eq:app-err-margin}. Take $x=\Eu(n)/\beta_{\mathrm{dedup}}(n)$ and $y=\Es(n)/\beta_{\mathrm{slot}}(n)$, so that $x/y=\bar u(n)/r(n)$. The margin~\eqref{eq:app-err-margin} is then $\bar u/r<\ell_-/\ell_+=(1-\varepsilon)/(1+\varepsilon)$, that is, $\ln r-\ln\bar u>\ln\frac{1+\varepsilon}{1-\varepsilon}$. The same argument with the kernels exchanged gives the other sign. The two cases together are~\eqref{eq:e3}.

\paragraph{Best draft length~\eqref{eq:e4}.}
Here $a\in[0,1)$, the true step time is $V(k)>0$ and the predicted one is $\hat V(k)$. They satisfy $(1-\delta)V(k)\le\hat V(k)\le(1+\delta)V(k)$ for every $k$, with $0\le\delta<1$. Over $\{0,\dots,\kmax\}$, the draft length $k^*$ maximizes $T_a=G_a/V$ and $\hat k^*$ maximizes $\hat T_a=G_a/\hat V$. Dividing $G_a(k)$ by the bounds on $\hat V(k)$ gives
\begin{equation}
\frac{T_a(k)}{1+\delta}\le\hat T_a(k)\le\frac{T_a(k)}{1-\delta}\qquad\text{for every }k
\label{eq:app-err-T}
\end{equation}
(\texttt{T\_bounds}). Since $\hat k^*$ maximizes $\hat T_a$,
\begin{equation}
\frac{1-\delta}{1+\delta}\,T_a(k^*)\le(1-\delta)\,\hat T_a(k^*)\le(1-\delta)\,\hat T_a(\hat k^*)\le T_a(\hat k^*)
\label{eq:app-err-regret}
\end{equation}
(\texttt{regret}). So the throughput lost, $1-T_a(\hat k^*)/T_a(k^*)$, is at most $1-(1-\delta)/(1+\delta)=2\delta/(1+\delta)$. If $\hat k^*\ne k^*$, \eqref{eq:app-err-regret} gives $T_a(k^*)\le\frac{1+\delta}{1-\delta}T_a(\hat k^*)$. The separation condition of~\eqref{eq:e4} is $\frac{1+\delta}{1-\delta}T_a(i)<T_a(k^*)$ for every $i\ne k^*$. For $i=\hat k^*$, this condition forbids the previous inequality. So under the separation condition, $\hat k^*=k^*$ (\texttt{argmax\_agree}, whose hypothesis is this condition). A common factor $c>0$ on every step time leaves the maximizer unchanged. The reason is that $G_a(i)/(cV(i))\le G_a(j)/(cV(j))$ exactly when $G_a(i)/V(i)\le G_a(j)/V(j)$ (\texttt{common\_factor}).

\paragraph{Best speedup~\eqref{eq:e5}.}
Here the assumptions are those of the previous paragraph. By~\eqref{eq:app-err-T}, $\hat T_a(\hat k^*)\ge\hat T_a(k^*)\ge T_a(k^*)/(1+\delta)$. Also $\hat T_a(\hat k^*)\le T_a(\hat k^*)/(1-\delta)\le T_a(k^*)/(1-\delta)$, so
\begin{equation}
\frac{T_a(k^*)}{1+\delta}\le\hat T_a(\hat k^*)\le\frac{T_a(k^*)}{1-\delta},\qquad \frac{T_a(0)}{1+\delta}\le\hat T_a(0)\le\frac{T_a(0)}{1-\delta}.
\label{eq:app-err-pair}
\end{equation}
Dividing the first pair by the second, in the direction that keeps each inequality, gives~\eqref{eq:e5} (\texttt{speedup\_bounds}).

\paragraph{From $\varepsilon$ to $\delta$.}
Here the bandwidths are the only errors of the predicted step time, and $\varepsilon<1/2$. By~\eqref{eq:e1}, the verify part of $\hat V(k)$ is off by at most $\varepsilon/(1-\varepsilon)$ of itself. An exact draft and host time $h(k)$ only lowers the relative error of the sum. So $\delta=\varepsilon/(1-\varepsilon)$ satisfies the assumption of~\eqref{eq:e4}. Then $1+\delta=1/(1-\varepsilon)$ and $1-\delta=(1-2\varepsilon)/(1-\varepsilon)$, so
\begin{equation}
\frac{1+\delta}{1-\delta}=\frac{1}{1-2\varepsilon},
\label{eq:app-err-delta}
\end{equation}
which for $\varepsilon=0.1$ is the factor 1.25 of Section~\ref{sec:spec-err}. This step is not in Lean.

\endgroup
\section{Engines and runs}
\label{sec:spec-tests}

\subsection{Kernel dispatch}
\label{sec:engines}

We follow each expert's matrix product, for each model and pass size $n$, to the GPU kernel that runs it, at pinned releases. We then count the requested bytes. The kernel fixes which bandwidth $\beta_p(n)$ of~\eqref{eq:tau} a pass of $n$ tokens runs at. On llama.cpp Metal, an expert product goes to a tiled matrix-matrix kernel only at 32 rows or more. Below that, a matrix-vector kernel multiplies each token's selected experts separately. MLX-LM switches to a tiled kernel only on passes of at least $4N/m$ tokens. The llama.cpp CUDA tiled kernel loads each selected expert once per tile on passes of 9 to 128 tokens. For Metal and MLX-LM, the step from each request pattern to its byte count is proved in Lean (\texttt{MoESpec.MetalClass} and \texttt{MoESpec.MlxClass}). The reading of the kernels from source is by hand. Table~\ref{tab:engines} gives the result.

\begin{table}[t]\centering\small

\caption{Kernel that runs the expert products, read from source. $n$ is the number of tokens verified in one pass, and the tests verify $n\le9$ tokens per step. ``Neither'': the requested bytes equal neither the union nor the per-slot count on every trace. Bytes are requested bytes. The source does not determine how many requests on-chip caches serve. The CUDA row at $2\le n\le8$ is the source reading of the matrix-vector kernel, which requests each slot's expert once. Profiled L2 requests of this kernel fall below $nmb$ when repeated experts sit in the same slot of every token. They fall as far as 0.14 of $nmb$, at $n=8$ and $U=8$. Two causes would both produce this: L1 hits under per-slot requests, and a pattern that requests each expert once per slot position. The hardware counters do not decide between them, so the row rests on the source. For gpt-oss, the factor $16/15$ counts the loads of the tiled kernel in the source reading. The factor $f_l=500/459$ of Table~\ref{tab:forks} instead counts the padded rows of each weight tensor, the byte unit of Appendix~\ref{app:dedup-evidence}.}
\label{tab:engines}
\renewcommand{\arraystretch}{1.15}
\begin{tabular}{p{0.14\textwidth}p{0.1\textwidth}p{0.29\textwidth}p{0.35\textwidth}}
\toprule
engine & tested sizes & other sizes & where the source decides \\
\midrule
llama.cpp Metal b11617 & per-slot & per-slot for $n\le31$; tiled (32 tokens per tile) for $32\le n\le512$, deduplicating only at $n=32$ for two models & dispatch test ``batch $\ge32$ rows'' in \texttt{ggml-metal-common.cpp}; matrix-vector kernel in \texttt{mul\_mv.metal}, one grid slice per token-slot pair \\
MLX-LM 0.32.0 & per-slot & per-slot for $n<4N/m$ (32 or 64); tiled (16 per tile), neither, above & dispatch branch on rows per expert in \texttt{mlx/backend/metal/quantized.cpp}; MoE call in \texttt{mlx\_lm/models/switch\_layers.py} \\
llama.cpp CUDA b11617 & upstream; run in Sec.~\ref{sec:dedup} & both at $n=1$; per-slot for $2\le n\le8$; deduplicating for $9\le n\le128$ (OLMoE, Qwen3), $16/15$ times union (gpt-oss) & batch limit 8 for the matrix-vector kernel in \texttt{ggml-cuda/mmvq.cuh}; dispatch in \texttt{ggml\_cuda\_mul\_mat\_id} (\texttt{ggml-cuda.cu}); tiles in \texttt{ggml-cuda/mmq.cu} \\
\bottomrule
\end{tabular}
\end{table}

\subsection{Models and prompts}
\label{app:spec-tests-inputs}

On llama.cpp, the models run as OLMoE-1B-7B-0125 in Q8\_0, Qwen3-30B-A3B in Q4\_K\_M and gpt-oss-20b in MXFP4 \citep{llamacpp}. On MLX-LM, OLMoE-1B-7B-0125 runs in 4-bit quantization. The oracle draft of Section~\ref{sec:dedup} marks the positions it corrupts by a seeded draw, and it costs essentially nothing. On the Apple engines, the realized acceptance is within 0.04 of $a$. On CUDA, where it was not logged per run, steps emit 2 to 5\% fewer tokens than $G_a(k)$ (Appendix~\ref{app:dedup-evidence}). Each draft length $k=0,\dots,8$ is run with each acceptance $a$ in an 11-value grid from 0.2 to 0.98. Each run generates 256 tokens on the six validation-set prompts of Section~\ref{sec:dedup}, at context depth $d=512$ and 4096. Each combination is run three times on the Apple engines and four times on CUDA. A run was discarded and redone if the machine was throttled, swapping, off AC power or running another inference process. Session load was not a discard reason. Other work on the M4 Max overlapped some whole-model runs. The run logs list these runs and are available on request. The DeepSeek-V2-Lite runs have no discard rule (Appendix~\ref{app:dedup-evidence}). Reuse $\hat\rho_l(n)$ is computed over every window of $n$ consecutive tokens in a fit document. For each window, it counts the last token's experts that an earlier token of the window also selected. The sum over windows is divided by $m$ times the number of windows. Its intervals come from 2000 bootstrap resamples of fit documents.

\subsection{Real draft model}
\label{sec:spec-tests-real}

Qwen3-30B-A3B (Q4\_K\_M) runs on llama.cpp Metal, drafted by Qwen3-0.6B (Q8\_0), with both models greedy. This engine runs the per-slot kernel on every pass of speculation (Table~\ref{tab:engines}). So each draft adds the expert bytes $mW$ of one more token. The runs use the 80 first-turn MT-Bench prompts \citep{mtbench2023} and generate 256 tokens at $k=0,\dots,8$. The run at $k=0$ uses the same program with a drafter that never drafts. The prompts were split into 42 for fitting an earlier time model and 38 held out from it. That earlier model is linear in the per-slot bytes, and this paper does not use it. We report the held-out prompts, of which 37 have a run at every draft length. Plain decoding is fastest on 33 of the 37. The median speedup over plain decoding is 0.877 at $k=1$ and 0.517 at $k=8$.

\section{Deduplication evidence}
\label{app:dedup-evidence}

\begin{table}[t]\centering\small

\caption{The deduplicating engines. The kernel each setting runs is read from source at $n=1,\dots,9$ verified tokens. At $n=1$, an engine runs the same kernel with deduplication off and on. $f_l$: bytes the deduplicating kernel reads per distinct expert, over the expert's bytes. On Qwen3 Q4\_K\_M, the tiled kernel reads each K-quant expert more than once. MMVQ and MMQ: llama.cpp CUDA's quantized matrix-vector and matrix-matrix kernels.}
\label{tab:forks}
\renewcommand{\arraystretch}{1.15}
\begin{tabular}{p{0.14\textwidth}p{0.16\textwidth}p{0.19\textwidth}p{0.19\textwidth}p{0.17\textwidth}}
\toprule
engine & switch & deduplication off & deduplication on & $f_l$, deduplicating \\
\midrule
llama.cpp Metal, modified & deduplicating kernel from 2 rows (32 upstream) & \texttt{mul\_mv\_id}, per-slot, $n\le9$ & \texttt{mul\_mm\_id}, deduplicating, $2\le n\le9$ & 1 (OLMoE Q8\_0, gpt-oss MXFP4); 2.14 or 2.33 (Qwen3 Q4\_K\_M, K-quant reload) \\
MLX-LM, modified & sort rows by expert, \texttt{gather\_qmm\_rhs} & \texttt{gather\_qmv}, per-slot, $n\le9$ & sorted rows, deduplicating, $2\le n\le9$ & 10/9 (16-row tile) \\
llama.cpp CUDA, modified & per-slot batch cap 8 to 1 & MMVQ, per-slot, $n\le8$; MMQ at $n=9$ & MMQ, deduplicating, $2\le n\le9$ & 1 (OLMoE); 500/459 (gpt-oss, padded unit) \\
\bottomrule
\end{tabular}
\end{table}

\paragraph{Engine changes.} The kernel each setting runs, its byte count and its factor $f_l$ are read from source at pinned commits. These are commits of the modified llama.cpp Metal and MLX-LM and of the unmodified CUDA build. On llama.cpp Metal, the minimum batch of the deduplicating kernel drops from 32 to 2. MLX-LM sorts rows by expert. On llama.cpp CUDA, an environment variable lowers the per-slot batch cap from 8 to 1. We checked each modified engine for equal output with deduplication off and on. On Metal, both settings give the same greedy tokens on OLMoE and gpt-oss, and Qwen3 diverges on one of three prompts at position 12. On CUDA, both settings give identical greedy tokens on both models and three prompts. There the verify argmax agrees on 132 of 132 rows (OLMoE) and 128 of 132 (gpt-oss, four near ties). On CUDA, the bytes read once per pass are the same with deduplication off and on, $D(n)=403062784+532480\,n$ on OLMoE.

\paragraph{Routing.} We measured reuse by teacher forcing on each modified engine's own greedy sequence. On OLMoE and Qwen3, it differs from the WikiText-103 values by $-0.018$ to $+0.004$ in $\rbar$. For gpt-oss, the RTX 4070 Ti SUPER routes the same expert set as the M4 Max on 0.947 of 11845 (row, layer) pairs. Its layer-mean $\hat\rho(n)$ is within 0.00083 at $n=2,\dots,9$. Neither measurement covers the greedy continuations at $d=4096$ (Section~\ref{sec:dedup-bytes}).

\begin{figure}[t]
\centering

\includegraphics[width=0.9\linewidth]{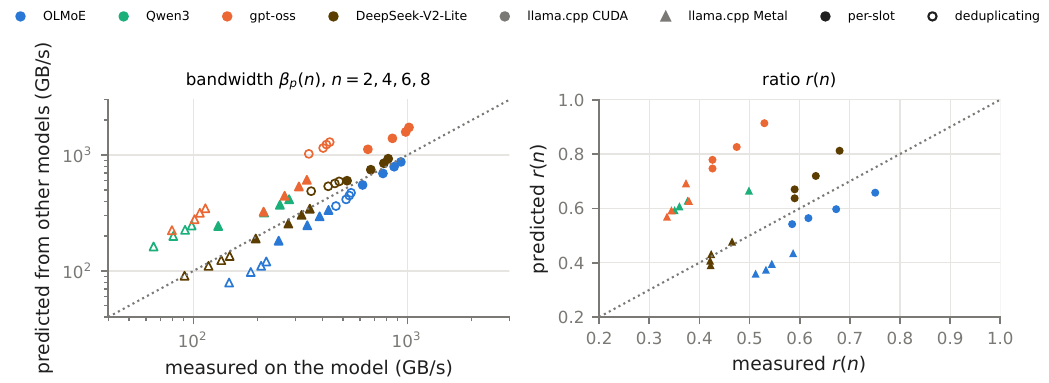}
\caption{Bandwidth of a held-out model, predicted from the other models of the same engine. It is plotted against the bandwidth measured on the expert operation of the held-out model, at $n=2,4,6,8$. Left: each kernel's bandwidth (filled: per-slot; open: deduplicating), log scale. Right: the ratio $r(n)$. Color: the held-out model; shape: engine (circle: llama.cpp CUDA, triangle: llama.cpp Metal). Dotted: equal.}
\label{fig:transfer}
\end{figure}

\paragraph{Deduplicating fit.}
Here, as for~\eqref{eq:reread}, $U$ is the number of distinct experts that a pass with chosen ids names. As there, $b$ is the bytes of one expert of the timed layer. We fit the median operation time of the deduplicating kernel per layer, by least squares over the passes with chosen ids at $n=2,\dots,9$, as
\begin{equation}
t_{\mathrm{dedup}}=t_{\mathrm{op}}+c_{\mathrm{g}}\,Ub+c_{\mathrm{t}}\,n,
\label{eq:app-dedup-fit}
\end{equation}
with its own fixed time $t_{\mathrm{op}}$, $c_{\mathrm{g}}$ the time per union byte and $c_{\mathrm{t}}$ the time per token. The benchmark of Section~\ref{sec:motivation} times one layer's expert operation through each engine's library (llama.cpp's ggml with the Metal or CUDA backend; MLX). It runs on the same model files, after the same 256 MiB cache flush. Its output was checked against a CPU reference on DeepSeek-V2-Lite (layers 1, 13 and 26, $n=1$, 4 and 9). The relative $L_2$ error is $1.0\times10^{-2}$ to $2.3\times10^{-2}$ on CUDA, where the GPU kernels quantize activations differently from the CPU. It is $9.5\times10^{-3}$ to $1.08\times10^{-2}$ on Metal. The chosen ids, the windows of recorded routing and every timing file are available on request. The prediction and the measured-time version read them through one script. This script also makes Figures~\ref{fig:beta} to~\ref{fig:deepseek} and~\ref{fig:transfer} and Tables~\ref{tab:beta}, \ref{tab:compare}, \ref{tab:blind} and~\ref{tab:perconfig}. It also computes the bandwidth error $\varepsilon$ of Section~\ref{sec:dedup-beta} from the fits on chosen ids.

\begin{table}[t]\centering\small

\caption{Per configuration, three quantities. Amplification: the factor $(\hat M^{\mathrm{e}}_{\mathrm{slot}}+\hat M^{\mathrm{e}}_{\mathrm{dedup}})/|S(n)|$ by which the saving bound~\eqref{eq:e2} turns a relative bandwidth error into a relative saving error. It is given as its range over $n=2,\dots,9$ ($n\le8$ on CUDA; $|S(n)|$ floored at 0.05 ms), and it is largest where $S(n)$ crosses 0. Implied $\varepsilon$: the bandwidth error that the prediction's largest saving error requires under~\eqref{eq:e2}, against the engine's $\varepsilon$. Verify-time error: the largest relative error of the verify time with deduplication on over $n=2,\dots,8$, for the prediction, the measured-time version and the calibrated version. Metal and CUDA are llama.cpp.}
\label{tab:perconfig}
\renewcommand{\arraystretch}{1.1}\setlength{\tabcolsep}{4pt}
\begin{tabular}{lll ccc ccc}
\toprule
& & & & \multicolumn{2}{c}{$\varepsilon$} & \multicolumn{3}{c}{verify-time error} \\
\cmidrule(lr){5-6}\cmidrule(lr){7-9}
engine & model & $d$ & amplification & implied & engine & prediction & measured-time & calibrated \\
\midrule
Metal & OLMoE & 512 & 6.0 to 79 & 0.014 & 0.031 & 19\% & 20\% & 6\% \\
Metal & OLMoE & 4096 & 6.2 to 86 & 0.028 & 0.031 & 24\% & 24\% & 6\% \\
Metal & Qwen3 & 512 & 3.9 to 7.8 & 0.020 & 0.031 & 21\% & 22\% & 6\% \\
Metal & Qwen3 & 4096 & 4.2 to 19 & 0.047 & 0.031 & 25\% & 31\% & 4\% \\
Metal & gpt-oss & 512 & 3.3 to 343 & 0.033 & 0.031 & 20\% & 15\% & 9\% \\
Metal & gpt-oss & 4096 & 3.8 to 87 & 0.126 & 0.031 & 6\% & 18\% & 8\% \\
MLX-LM & OLMoE & 512 & 3.2 to 3.8 & 0.108 & 0.148 & 38\% & 30\% & 26\% \\
MLX-LM & OLMoE & 4096 & 3.3 to 3.8 & 0.107 & 0.148 & 41\% & 31\% & 17\% \\
CUDA & OLMoE & 512 & 15.4 to 139 & 0.030 & 0.054 & 3\% & 9\% & 3\% \\
CUDA & OLMoE & 4096 & 15.2 to 146 & 0.032 & 0.054 & 5\% & 5\% & 2\% \\
CUDA & gpt-oss & 512 & 7.5 to 223 & 0.022 & 0.054 & 9\% & 16\% & 6\% \\
CUDA & gpt-oss & 4096 & 6.0 to 43 & 0.033 & 0.054 & 21\% & 4\% & 1\% \\
\bottomrule
\end{tabular}
\end{table}

\paragraph{Speculative runs.} The measured speculative runs on CUDA differ from $G_a(k)/(\tau_{\mathrm{dedup}}(k+1)+h(k))$ with the measured $\tau_{\mathrm{dedup}}$ and $h$. Per configuration, the median factor between them is 0.93 to 1.04, over a range of 0.74 to 1.18. Two terms make up this difference. First, steps emit 2 to 5\% fewer tokens than $G_a(k)$. On steps that run to their full length, the shortfall is 1.6 to 2.7\%. This part comes from the sampling of the seeded flags that mark drafts wrong. The rest comes from steps cut short where the argmax of the batched verify pass differs from the batch-one greedy token that the oracle draft proposes. This happens on up to 25.8\% of steps, at $k=8$, $a=0.98$ on gpt-oss at $d=4096$. That term belongs to the oracle draft, and a real draft model checked against the batched pass has none. Second, the time per step beyond the verify pass is 0.58 to 0.61 of the measured $h(k)$. The verify time departs from its fit by $-7$ to $+9\%$. Against the speedups of the measured speculative runs, the speedup bound~\eqref{eq:e5} therefore holds at 9 of 11 acceptance values on CUDA gpt-oss at $d=4096$. It holds at all values elsewhere. The two misses come from these two terms.

\paragraph{Scoring rules.} Table~\ref{tab:blind-criteria} lists the six scoring rules written before the whole-model runs of Section~\ref{sec:dedup-blind}, with the value each measurement gives. The scoring script decides each rule at each depth. The measured-time version meets the rules only if all six hold at both depths. The same script, byte for byte, scored both engines. On Metal, the whole-model runs at $d=512$ show a slow GPU state in some runs. In each of the four whole-model runs at $k=0$, the first prompt runs at 6.7 to 7.1 ms per step and the others at 7.5 to 9.4 ms. With a 90 s rest after every 10 minutes of runs, the state persists. The system reports no thermal warning at any of the 37 rests. At $d=4096$, the ratio of the slowest to the fastest prompt's median step time has a median of 1.03. It exceeds 1.08 in 2 of 94 runs at acceptance 1. The rules exclude no runs, so these runs are scored.

\begin{table}[t]\centering\small

\caption{Scoring rules for DeepSeek-V2-Lite and the measured values, measured-time version. The speedup rule uses $\delta$, the largest relative error of the predicted step time. The crossover rule is scored over $n=2,\dots,9$, and a measured crossover at $n=9$ counts as within one of none. On CUDA, the deduplicating kernel runs at $n=9$ with deduplication off and on. So Table~\ref{tab:blind}, like Table~\ref{tab:compare}, reads the CUDA crossover over $n\le8$.}
\label{tab:blind-criteria}
\renewcommand{\arraystretch}{1.12}
\begin{tabular}{p{0.36\textwidth}cccc}
\toprule
rule & CUDA 512 & CUDA 4096 & Metal 512 & Metal 4096 \\
\midrule
crossover within one & $-$/$-$, holds & $-$/9, holds & $-$/$-$, holds & $-$/$-$, holds \\
$\hat k^*$ within one at $\ge9$ of 11 values & 11, holds & 11, holds & 9, holds & 6, fails \\
loss at most 6\% at every value & 2.9\%, holds & 4.1\%, holds & 20.1\%, fails & 22.5\%, fails \\
sign of $S(n)$ where the interval excludes 0 & 7 of 7, holds & 7 of 7, holds & 8 of 8, holds & 8 of 8, holds \\
verify time within 15\%, both kernels & 9.9\%, holds & 10.1\%, holds & 34.0\%, fails & 28.4\%, fails \\
best speedup inside~\eqref{eq:e5} & $\delta=0.072$, holds & 0.077, holds & 0.27, holds & 0.25, holds \\
\bottomrule
\end{tabular}
\end{table}

\paragraph{Transfer.}\label{sec:dedup-transfer} Characterizing the bandwidth once per engine, instead of measuring it per model, fails (Figure~\ref{fig:transfer}). We hold out one model at a time on CUDA (three models) and Metal (four). We fit $\log\beta_p(n)$ on the other models as linear in $\log n$ and the log of the expert bytes, a form fixed before any held-out prediction. This fit misses the held-out bandwidth by 9 to 206\% (median 23\% on CUDA, 47\% on Metal). It misses the ratio $r$ by 2 to 30\% on OLMoE and DeepSeek-V2-Lite and by 33 to 86\% on gpt-oss and Qwen3. The expert sizes of gpt-oss and Qwen3 lie outside the range of the other models. Near the crossover, this error exceeds the margin of~\eqref{eq:e3}. The transferred bandwidth places the crossover within one in 1 of 6 CUDA configurations and 2 of 8 Metal configurations. With each model's own measurement, the counts are 6 of 6 and 7 of 8. The quantization format (Q8\_0, Q4\_K\_M, MXFP4) and the model are confounded, and MLX-LM has one model.

\end{document}